\documentclass[10pt,journal,compsoc]{IEEEtran}
\ifCLASSINFOpdf
\else
\fi
\usepackage[table,xcdraw]{xcolor} 
\usepackage{amssymb} 
\usepackage[linguistics]{forest} 
\usepackage{booktabs}
\usepackage{graphicx} 
\usepackage[numbers,sort&compress]{natbib}
\newcommand{\etal}{\textit{et al.}}
\usepackage{multirow} 
\usepackage{hyperref}
\hypersetup{
    colorlinks=true,
    linkcolor=blue,
    filecolor=magenta,      
    urlcolor=cyan,
    pdftitle={Overleaf Example},
    pdfpagemode=FullScreen,
    }

\newcommand{\zty}[1]{\textcolor{black}{#1}}

\begin{document}
%
\title{\zty{Graph-based Knowledge Distillation: A survey and experimental evaluation}}
%
%
%
%

\author{Jing Liu,
        Tongya Zheng,
        Guanzheng Zhang,
        Qinfen Hao
\IEEEcompsocitemizethanks{\IEEEcompsocthanksitem Jing Liu is with the SKLP, Institute of Computing Technology, Chinese Academy of Sciences, Beijing, China, 100190.\protect\\
E-mail: liujing18z@ict.ac.cn
\IEEEcompsocthanksitem Tongya Zheng is with College of Computer Science and Technology, Zhejiang University, Hangzhou, China, 310027.\protect\\
E-mail: tyzheng@zju.edu.cn
\IEEEcompsocthanksitem Guanzheng Zhang is with College of Information Engineering, China University of Geosciences (Beijing), Beijing, China, 100083.\protect\\
\IEEEcompsocthanksitem Qinfen Hao is with the SKLP, Institute of Computing Technology, Chinese Academy of Sciences, Beijing, China, 100190. Correspondence to: Qinfen Hao.\protect\\
E-mail: haoqinfen@ict.ac.cn
}
\thanks{Manuscript received xx xx, xx; revised xx xx, xx.}}

%
%

\markboth{Journal of IEEE}%
{Shell \MakeLowercase{\textit{et al.}}: Bare Advanced Demo of IEEEtran.cls for IEEE Computer Society Journals}
%



\IEEEtitleabstractindextext{%
\begin{abstract}
Graph data, such as citation networks, social networks, and transportation networks, are prevalent in the real world. Graph neural networks (GNNs) have gained widespread attention for their robust expressiveness and exceptional performance in various graph analysis applications. However, the efficacy of GNNs is heavily reliant on sufficient data labels and complex network models, with the former being challenging to obtain and the latter requiring expensive computational resources. To address the labeled data scarcity and high complexity of GNNs, Knowledge Distillation (KD) has been introduced to enhance existing GNNs. This technique involves transferring the soft-label supervision of the large teacher model to the small student model while maintaining prediction performance.
Transferring the KD technique to graph data and graph-based knowledge is a major challenge. This survey offers a comprehensive overview of Graph-based Knowledge Distillation methods, systematically categorizing and summarizing them while discussing their limitations and future directions.
This paper first introduces the background of graph and KD. It then provides a comprehensive summary of three types of Graph-based Knowledge Distillation methods, namely Graph-based Knowledge Distillation for deep neural networks (DKD), Graph-based Knowledge Distillation for GNNs (GKD), and Self-Knowledge Distillation based Graph-based Knowledge Distillation (SKD). Each type of method is further divided into knowledge distillation methods based on the output layer, middle layer, and constructed graph. Subsequently, various graph-based knowledge distillation algorithms' ideas are analyzed and compared, concluding with the advantages and disadvantages of each algorithm supported by experimental results. In addition, the applications of graph-based knowledge distillation in computer vision, natural language processing, recommendation systems, and other fields are listed. Finally, the development of graph-based knowledge distillation is summarized and prospectively discussed. We have also released related resources at \url{https://github.com/liujing1023/Graph-based-Knowledge-Distillation}.

\end{abstract}

\begin{IEEEkeywords}
Graph, Graph Neural Networks (GNNs), Knowledge Distillation (KD), Self-Knowledge Distillation (Self-KD).
\end{IEEEkeywords}}

\maketitle

\IEEEdisplaynontitleabstractindextext

%
\IEEEpeerreviewmaketitle


\section{Introduction}
\IEEEPARstart{G}{raph} data~\cite{Graphdata}, which represents the relationship between objects, is an important data type used in various real-world scenarios such as user recommendation~\cite{UserRS}, drug discovery~\cite{Drugdiscovery}, traffic forecasting~\cite{Trafficforecast}, point cloud classification~\cite{Pointcloud}, and chip design~\cite{Chipdesign}. Unlike structured data in Euclidean space, graph data has a complex structure and contains rich information. To learn vectorized representations with sufficient information from complex graphs, researchers are applying deep learning methods to graphs. Drawing on the idea of convolutional neural networks (CNNs)~\cite{CNN}, graph neural networks (GNNs)~\cite{GNN} have been proposed, and they have been effectively applied in tasks such as node classification~\cite{Nodeclf}, link prediction~\cite{LinkPred}, and graph classification~\cite{Graphclf}.

As convolution operators improve and large-scale graphs become more prevalent, researchers are exploring ways to train accurate and efficient graph convolution neural networks (GNNs). One approach is to train deeper networks to improve generalization. However, GNNs are semi-supervised and rely on high-quality labeled data and complex models, which are difficult to obtain and computationally expensive.

To address the challenges of sparse data labeling and high model complexity in GNNs, knowledge distillation (KD)\cite{KD} is introduced to graph analysis. KD is a "Teacher-Student" (T-S) network training method that transfers soft label knowledge learned by the T network with strong learning ability to the S network with small parameters and weak learning ability to improve its performance. This achieves model compression and results in an effect similar to that of T. KD is widely applied in academia and industry, such as in computer vision\cite{KDincv}, speech recognition~\cite{Kdinspeech}, natural language processing~\cite{Kdinnlp}, etc., due to its simplicity and effectiveness.

Recently, the potential of applying the T-S knowledge distillation framework to GNNs has been demonstrated. Researchers have designed knowledge distillation algorithms for graph data or directly for GNNs by combining KD with GNNs, motivated by the success of KD on CNNs. The first work in this area was LSP~\cite{LSP}, which applied KD to GCN~\cite{GCN} by proposing the local structure retention module to distill local graph structure knowledge from deep GCN teacher models into shallow GCN student models with fewer parameters. Subsequently, other graph-based knowledge distillation methods have been proposed. Although KD has shown promising progress in GNNs, the existing methods have mainly focused on CNNs with structured grid data as input, and there are few studies on GNNs with irregular data processing capabilities. Additionally, there is a lack of comprehensive review of \zty{Graph-based Knowledge Distillation} research. This paper aims to fill this gap by providing a systematic review of existing knowledge distillation work on graphs.

To summarize, the contributions of this work are listed as follows:

In summary, this work makes the following contributions:
\begin{itemize}
    \item We present the first comprehensive review of \zty{Graph-based Knowledge Distillation} by covering more than 100 papers, filling the gaps in this field. The survey includes problem definition, theoretical analysis, method classification, experimental comparison, and application and prospects.
    \item We use hierarchical classification to systematically summarize and analyze the latest progress in \zty{Graph-based Knowledge Distillation} methods, providing insights for each type of method (refer to Fig.~\ref{fig:taxonomy}).
    \item We conduct extensive experiments to compare the distillation effect of each type of knowledge distillation method and provides in-depth analysis.
    \item We discuss the challenges of existing \zty{Graph-based Knowledge Distillation}, propose potential research directions and trends for the future, and provide insightful guidance for researchers in the fields of GNNs and KD.
    \item We establish an open-source code library\footnote{https://github.com/liujing1023/Graph-based-Knowledge-Distillation\label{library}} for \zty{Graph-based Knowledge Distillation} research, which serves as a valuable reference for this research field.
\end{itemize}

The paper is organized as follows. Section~\ref{sec:Background} motivates the review of \zty{Graph-based Knowledge Distillation}, discusses its latest progress, and its relation to existing research areas. Section~\ref{sec:Theory} formalizes each type of \zty{Graph-based Knowledge Distillation} method and presents the related theory. Section~\ref{sec:Methodology} summarizes the method categorization, including \zty{Graph-based Knowledge Distillation} for deep neural networks (DKD), \zty{Graph-based Knowledge Distillation} for graph neural networks (GKD), and \zty{Self-Knowledge Distillation} based \zty{Graph-based Knowledge Distillation} (SKD), and further subdivides them into methods based on output layer, middle layer, and constructed graph. Section~\ref{sec:Experiment} compares and analyzes classical \zty{Graph-based Knowledge Distillation} algorithms. Section~\ref{sec:Applications} lists the application of \zty{Graph-based Knowledge Distillation} methods in CV, NLP, RS, and other scenarios. Section~\ref{sec:Discussion} prospects the future research direction of \zty{Graph-based Knowledge Distillation}. Finally, Section~\ref{sec:Conclusion} concludes the paper. We also establish an open-source code library\textsuperscript{\ref{library}} for \zty{Graph-based Knowledge Distillation} research.

\definecolor{level1}{HTML}{C5E0B4}
\definecolor{level2}{HTML}{DEEBF7}
\definecolor{level3}{HTML}{FFD8D8}
\definecolor{level4}{HTML}{FFF2CC}
\definecolor{level5}{HTML}{cafcda}

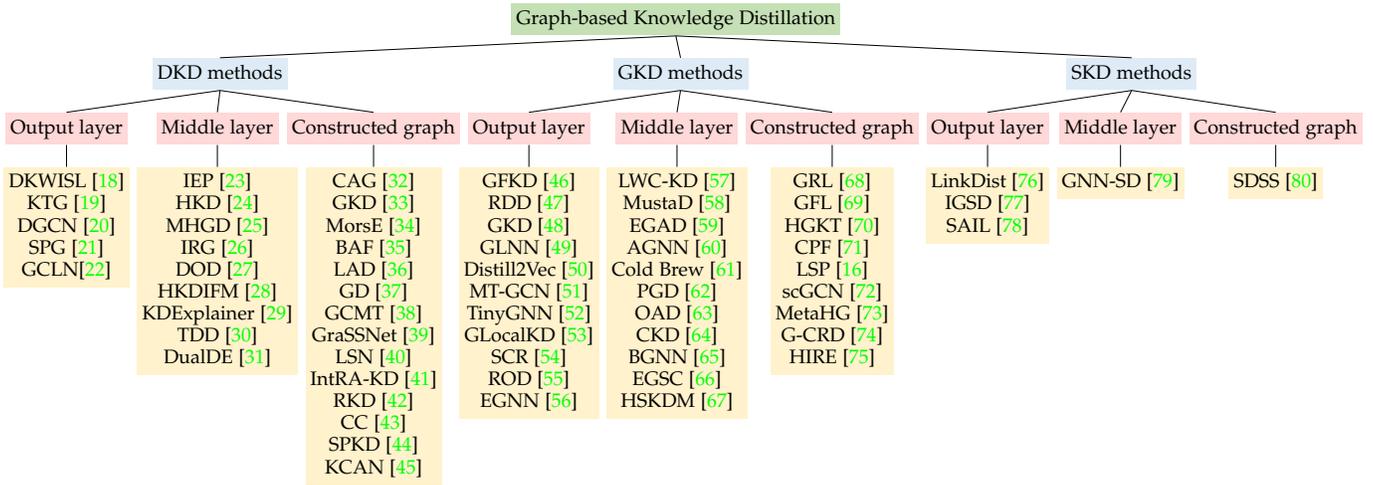
\begin{figure*}
    \centering
    \medskip
    \footnotesize
    
    \resizebox{\linewidth}{!}{
     \begin{forest}
        for tree={s sep=1mm, inner sep=2, l=1}
        [
        {Graph-based Knowledge Distillation },fill=level1
            [DKD methods,fill=level2 
            [Output layer,fill=level3 
            [{DKWISL~\cite{DKWISL}}\\KTG~\cite{KTG}\\DGCN~\cite{DGCN}\\SPG~\cite{SPG}\\GCLN\cite{GCLN},fill=level4]]
            [Middle layer,fill=level3
            [IEP~\cite{IEP}\\HKD~\cite{HKD}\\MHGD~\cite{MHGD}\\IRG~\cite{IRG}\\DOD~\cite{DOD}\\HKDIFM~\cite{HKDIFM}\\KDExplainer~\cite{KDExplainer}\\TDD~\cite{TDD}\\DualDE~\cite{DualDE} ,fill=level4]]
            [Constructed graph,fill=level3 
            [CAG~\cite{CAG}\\GKD~\cite{GKD}\\MorsE~\cite{MorsE}\\BAF~\cite{BAF}\\LAD~\cite{LAD}\\GD~\cite{GD}\\GCMT~\cite{GCMT}\\GraSSNet~\cite{GraSSNet}\\LSN~\cite{LSN}\\IntRA-KD~\cite{IntRA-KD}\\RKD~\cite{RKD}\\CC~\cite{CC}\\SPKD~\cite{SPKD}\\KCAN~\cite{KCAN},fill=level4]
            ]]
            [{GKD methods},fill=level2 
            [{Output layer},fill=level3 
            [GFKD~\cite{GFKD}\\RDD~\cite{RDD}\\GKD~\cite{GKD2}\\GLNN~\cite{GLNN}\\Distill2Vec~\cite{Distill2Vec}\\MT-GCN~\cite{MT-GCN}\\TinyGNN~\cite{TinyGNN}\\GLocalKD~\cite{GLocalKD}\\SCR~\cite{SCR}\\ROD~\cite{ROD}\\EGNN~\cite{EGNN},fill=level4
            ]
            ]
            [{Middle layer},fill=level3 
            [LWC-KD~\cite{LWC-KD}\\MustaD~\cite{MustaD}\\EGAD~\cite{EGAD}\\AGNN~\cite{AGNN}\\Cold Brew~\cite{ColdBrew}\\PGD~\cite{PGD}\\OAD~\cite{OAD}\\CKD~\cite{CKD}\\BGNN~\cite{BGNN}\\EGSC~\cite{EGSC}\\HSKDM~\cite{HSKDM},fill=level4]
            ]
            [{Constructed graph},fill=level3 
            [GRL~\cite{GRL}\\GFL~\cite{GFL}\\HGKT~\cite{HGKT}\\CPF~\cite{CPF}\\LSP~\cite{LSP}\\scGCN~\cite{scGCN}\\MetaHG~\cite{MetaHG}\\G-CRD~\cite{G-CRD}\\HIRE~\cite{HIRE},fill=level4]
            ]]
            [{SKD methods},fill=level2 
            [Output layer,fill=level3
            [LinkDist~\cite{LinkDist}\\IGSD~\cite{IGSD}\\SAIL~\cite{SAIL},fill=level4]]
            [Middle layer,fill=level3
            [GNN-SD~\cite{GNN-SD},fill=level4]
            ]
            [Constructed graph,fill=level3
            [SDSS~\cite{SDSS},fill=level4]
            ]]
        ]
    \end{forest}
    } %
    \caption{Taxonomy of Graph-based Knowledge Distillation methods. }
    \label{fig:taxonomy}
\end{figure*}

\section{Background}\label{sec:Background}
Before introducing the \zty{Graph-based Knowledge Distillation} \zty{methods}, this section first gives the basic concepts and symbolic definitions involved in the \zty{Graph-based Knowledge Distillation} technology, and then briefly explains the development and division criteria of the \zty{Graph-based Knowledge Distillation} method.

\subsection{\zty{Graph Neural Networks}}
It is no doubt that deep learning has achieved great success in structured data in \zty{Euclidean} space, but real-life data is naturally modeled as unstructured data such as graphs. As a common data structure, graph can be represented as a set of vertices $V$ and edges $E$, denoted $G = (V, E)$. Graph is widely used in graph analysis because of its powerful expressive. For example, in the e-commerce recommendation field~\cite{MEIRec}, a graph-based learning system is needed to realize highly accurate recommendations by utilizing the interaction between users and items. The complexity of graphs poses major challenges to existing machine learning algorithms: each graph has a different size, nodes are disordered, and each node in a graph has a different number of neighbor nodes, which makes the conventional convolution operation in deep learning unable to be directly applied to graphs. Recently, inspired by the convolution idea of CNNs, researchers have tried to apply deep learning methods to graph analysis. Since then, GNNs have become a widely used tool for graph analysis due to their excellent performance and interpretability.

Recently, the research enthusiasm for GNNs in deep learning field has been increasing, and it has become a research hotspot in various fields. GNNs have made new breakthroughs in biochemistry~\cite{Chemical}, physical modeling~\cite{Physics}, knowledge graph~\cite{KG}, and circuit design~\cite{Circuit}.  With the development of the graph neural network model, GNNs can be divided into the spectral method and spatial methods.

On the one hand, the spectral-based method introduces filters to define graph convolution from the perspective of graph signal processing. In 2013, Bruna \etal~\cite{SpectralGNN} introduced the concept of frequency-domain convolutional operation into GNNs based on spectral theory~\cite{Spectraltheory}. Hence, the first spectral method named Spectral CNN was proposed for the first time. Subsequently, the spectral-based graph convolutional network method has been further improved and expanded~\cite{ChebyNet, AGCN, zhuang2018dual, xugraph}. For instance, ChebyNet~\cite{ChebyNet} parameterized kernel convolution using the matrix form of Chebyshev polynomial to greatly reduce the parameters and computational complexity of Spectral CNN, thus making the spectral method practical. Nevertheless, the spectral method usually needs to process the entire graph at the same time when calculating and needs to bear the high time complexity of matrix decomposition, which is difficult to parallel or extend to large graphs. Therefore, graph convolutional networks based on spatial domains begin to develop rapidly.

On the other hand, the spatial-based method directly performs convolution operations on a graph and represents the graph convolution as aggregating feature information from the neighborhood. As a representative work of spatial methods, GCN~\cite{GCN} further simplifies graph convolution in the spectral domain by using first-order approximation, which enables graph convolution operations to be carried out in the spatial domain and greatly improves the computational efficiency of graph convolution models. Moreover, to speed up the training of graph neural networks, GNNs can also be combined with sampling strategies, including SAGE~\cite{SAGE}, FastGCN~\cite{FastGCN}, LADIES~\cite{LADIES}, etc., to efficient computation by limiting the computation to a batch of nodes rather than the entire graph (alleviate problems such as training time and memory requirements, etc.). Subsequently, to make GCN more powerful, more spatial-based GNNs~\cite{GAT, atwood2016diffusion, MPNN,wang2018non, RGCN} were proposed, and remarkable results were achieved in a variety of graph-related tasks. Given the advantages of a high degree of freedom, excellent computability, and high reasoning efficiency, the spatial-based method has received extensive attention and development. Additionally, many scholars have combed and summarized GNNs from different perspectives (such as methods, applications, etc.). For details, please refer to the review~\cite{zhou2020graph, wu2020comprehensive, zhang2020deep, skarding2021foundations, yang2020heterogeneous, sun2018adversarial, wu2023graph, nazir2021survey, lopera2021survey, wu2022graph, lamb2020graph}.  Due to its high degree of freedom, good computability, and high reasoning efficiency, the spatial-based method has been widely concerned and developed.

Obviously, GNNs have been proven to be a powerful non-grid data model, but the original GNN still has some limitations. There are two main points: (1) the existing GNNs are mostly semi-supervised learning, which makes their performance depend heavily on high-quality labeling data; (2) with the development of graph scale, the design of existing graph models is becoming more and more complex, bringing certain challenges to graph model calculation and graph storage. The successful application of KD in computer vision provides a feasible scheme for the above two challenges. Section~\ref{sec:kdindnn} will briefly review the development of KD in deep learning.

\subsection{\zty{Knowledge Distillation}}\label{sec:kdindnn}
Knowledge distillation~\cite{KD} was originally proposed for model compression. Unlike pruning and quantification in model compression, knowledge distillation (KD) uses the T-S framework to pre-train a large teacher model to distill to obtain a lightweight student model, enhancing the generalization ability of the student model and achieving better performance and higher precision. Through distillation, "knowledge" (soft label supervised information) in the teacher model is transferred to the student model. In this way, student models can reduce the complexity of time and space, which can also learn soft label information (containing inter-category information) that is not available in the one-hot label without losing the quality of the prediction. Generally, KD can be divided into two technical routes in accordance with the different ways of knowledge transfer.

The first is response-based distillation, which is closely related to label smoothing~\cite{Labelsmooth}, using the output probability of the teacher model as smoothing labels to train students. ~\cite{KD} is the pioneering work of knowledge distillation, proposed by Hinton in 2015, which was first proposed to transfer the output probability by the softmax layer of the teacher model to the student model as a "soft-target" to improve the performance of the student model. To learn the feedback information in the student network, DML~\cite{DML} proposes the strategy of deep mutual learning, allowing a group of students to train simultaneously on the network and realize mutual learning and progress through the supervision of real labels and the learning experience of peer network output results. BAN~\cite{BAN} uses an integrated approach to train the student model so that its network structure is the same as that of the teacher model, which significantly outperforms the teacher model in computer vision and language modeling downstream tasks.

Another type of knowledge distillation is the feature-based distillation method, in which the semantic information contained in the middle layer feature representation in the teacher network structure as knowledge transfer to the student model.   FitNet~\cite{FitNet} is the first classical work to adopt this method, leveraging the output of the teacher network and the feature embedding of the middle layer as supervision information to extend KD and realize the problem of deep model network compression.   Feature-based methods have become mainstream, including attention mechanism~\cite{AT}, probability distribution matching~\cite{FT, Alp-kd}, etc.   After that, new relation-based distillation methods have been derived from feature-based distillation methods~\cite{yim2017gift, passalis2020probabilistic, DKWISL, KTG, DGCN, SPG, GCLN}, but they are all designed to better distill the feature-based knowledge from the teacher to the student.

Regardless of the distillation strategy, most of these methods are designed for CNNs with grid data as input. Fortunately, methods have emerged to design knowledge distillation on graphs and GNNs recently. In the following, we will briefly summarize the \zty{Graph-based Knowledge Distillation} method for designing distillation algorithms on graphs.

\section{\zty{\zty{Graph-based Knowledge Distillation}}}\label{sec:Theory}

With the development of knowledge distillation techniques, distillation methods that only distill information from a single sample are no longer applicable because they provide limited information.  To extract rich correlation information between different data samples, the relation-based knowledge distillation method~\cite{yim2017gift, passalis2020probabilistic, DKWISL, KTG, DGCN, SPG, GCLN} is proposed, which fully mines the structural feature knowledge between samples in teacher networks by implicitly/explicitly constructing the relationship graph between samples.  As a powerful unstructured modeling tool, GNNs can directly model graph data.  Therefore, distillation with GNNs can easily extract and transmit graph topology knowledge and semantic supervision information between samples.  Therefore, using graph neural networks for distillation can easily realize the extraction and transmission of graph topological structure knowledge and semantic supervision information between samples.  Thus, we refer to the relation-based knowledge distillation method based on deep neural networks (DNNs) and the distillation method based on GNNs as the \zty{Graph-based Knowledge Distillation} method.  \zty{Graph-based Knowledge Distillation} aims at distilling directly/indirectly constructed sample relationship semantic information in the teacher model into the student model in order to obtain more general, richer, and more sufficient knowledge.

Despite GNN being a powerful architecture with excellent performance in modeling unstructured data, its superior performance depends on high-quality tag data and complex network models. Nevertheless, label acquisition is difficult to obtain, and computing resources are costly. Hence, in the face of sparse data labels and the high complexity of model computation in GNNs, how to design smaller and faster networks with guaranteed performance has become the focus of research. Based on this idea, various methods have emerged to graph design knowledge distillation algorithms. Meanwhile, with the excellent performance of KD in graph analysis tasks, the study of \zty{Graph-based Knowledge Distillation} has been widely concerned.

This paper presents a hierarchical classification method for Graph-based Knowledge Distillation, dividing it into Graph-based Knowledge Distillation is divided into Graph-based Knowledge Distillation for deep neural networks (DKD), Graph-based Knowledge Distillation for graph neural networks (GKD), and Self-Knowledge Distillation based Graph-based Knowledge Distillation (SKD). The classifications are further based on the distillation positions, including the output layer, the middle layer, and the constructed graph. Fig.\ref{fig:taxonomy} shows the specific classifications and representative methods. SKD is focused on the Self-Knowledge Distillation distillation method in GNN models due to the recent attention on the combination of Self-Knowledge Distillation and GNNs. We analyze DKD, GKD, and SKD methods and summarize the common symbols and meanings in TABLE\ref{tab:symbol}.

\begin{table}[htbp]
  \caption{Mathematical notations used in this paper.}
  \centering
  \begin{tabular}{l|p{0.7\linewidth}}
    \toprule
    Notation & Description \\ 
    \hline
    $X, Y$ & the dataset and label\\
    \hline
    $n$ & the number of samples\\
    \hline
    $T, S$ & the teacher and student model\\
    \hline
    $W_t, W_s$ & the model parameters of $T$/$S$ \\
    \hline
    $z_t,z_s$ & the probability output of $T$/$S$ \\
    \hline
    $p_t,p_s$ & the probability distribution of $T$/$S$ \\
    \hline
    $\tau $ & the temperature scaling factor\\
    \hline
    $\mathcal{L } _{CE} $ & the cross-entropy loss\\
    \hline
    $\mathcal{L } _{KD} $ & the distillation loss\\
    \hline
    $\mathcal{L } _{D} $ & the graph distillation loss for DKD\\
    \hline
    $\mathcal{L } _{G} $ & the graph distillation loss for GKD\\
    \hline
    $\mathcal{L } _{Self} $ & the graph distillation loss for SKD\\
    \hline
    $G, V, E$ & a graph, the node-set in $G$ , the edge-set in $G$\\
    \hline
    $\mathcal{L }$ & the total number of convolution layers\\
    \hline
    $h^{l} $ & the node embedding at $l$-th layer \\
    \hline
    $v, u$ & the target node, the neighbor node of $v$ \\
    \hline
    $x_t,x_s$ & a node of $T$/$S$  \\
     \hline
    $\alpha $ & the hyperparameter of graph distillation loss \\
    \hline
     $D_{D},D_{G},D_{Self} $ & the distance measure function\\
    \hline
     $\psi_{t}$ & the sample similarity function\\
    \hline
     $S$ & the node similarity function\\
    \hline
     $G_t,G_s$ & the internode relationship constructor of $T$/$S$\\
    \bottomrule
  \end{tabular}
  \label{tab:symbol}
\end{table}

\subsection{\zty{\zty{Graph-based Knowledge Distillation} for Deep Neural Networks}}
Firstly, for simplicity, the well-performed teacher model with parameter $W_t$ is denoted as $T$. Similarly, the student network model with parameter $W_s$ is indicated as $S$. The convolutional neural network input dataset is represented as $X=\left\{ {x_{1},x_{2},\cdots,x_{n}} \right\}$, and the corresponding label is $Y=\left\{ {y_{1},y_{2},\cdots,y_{n}} \right\}$, where $n$ indicates the number of samples in the dataset. Since the DNNs can be regarded as mapping functions superimposed by multiple nonlinear layers, the non-normalized probability outputs of teachers and students are presented as $z_{t} = \varphi\left( {X;W_{t}} \right)$ and $z_{s} = \varphi\left( {X;W_{s}} \right)$, where $\varphi$ is the mapping function. $p_{t} = softmax\left( z_{t} \right)$ and $p_{t} = softmax\left( z_{t} \right)$ represent the final predicted probabilities of teachers and students, respectively.

Knowledge distillation was first proposed by Hinton \etal~\cite{KD}, aiming to transfer knowledge hidden in a large network (teacher model, $t$) to a lightweight network (student model, $s$) so that the student model can achieve better performance. The basic idea of knowledge distillation is to soften the class probability distribution of the softmax output layer by temperature $\tau$ to obtain the soft target:
\begin{equation} 
  p^{\tau} = \frac{exp\left( \frac{z_{i}}{\tau} \right)}{\sum_{j}{exp\left( \frac{z_{j}}{\tau} \right)}}~,
  \label{eq:kd-p}
\end{equation}
where $\tau$ represents the temperature scaling coefficient and is used to soften the output of the teacher model. The larger $\tau$ is, the smoother the output probability will be.

Hinton \etal also found that guiding the student model together with the soft target and the ground truth during the training process will further improve the learning effect, specifically by weighting the loss function of the two parts. Thus, the loss of knowledge distillation can be expressed as:
\begin{equation} 
  \mathcal{L }_{KD} = L_{CE}\left( {p_{s},y} \right) + \alpha\tau^{2}KL\left( {p_{s}^{\tau},p_{t}^{\tau}} \right),
  \label{eq:kd-loss}
\end{equation}
where the term $L_{CE}$ is the traditional cross-entropy loss. That is the cross entropy between the predicted output $p_{s}$ of the student model and the ground truth $y$. The second term is the cross-entropy between the predicted output of the student model after $\tau$ smooth $p_{s}^{\tau}$ and the teacher model after $\tau$ smooth output $p_{t}^{\tau}$. $\alpha$ is the hyperparameter that adjusts the ratio of the two loss functions. KL is the Kullback-Leibler divergence.

\begin{figure}[htbp]
    \centering
  \includegraphics[width=0.5\textwidth]{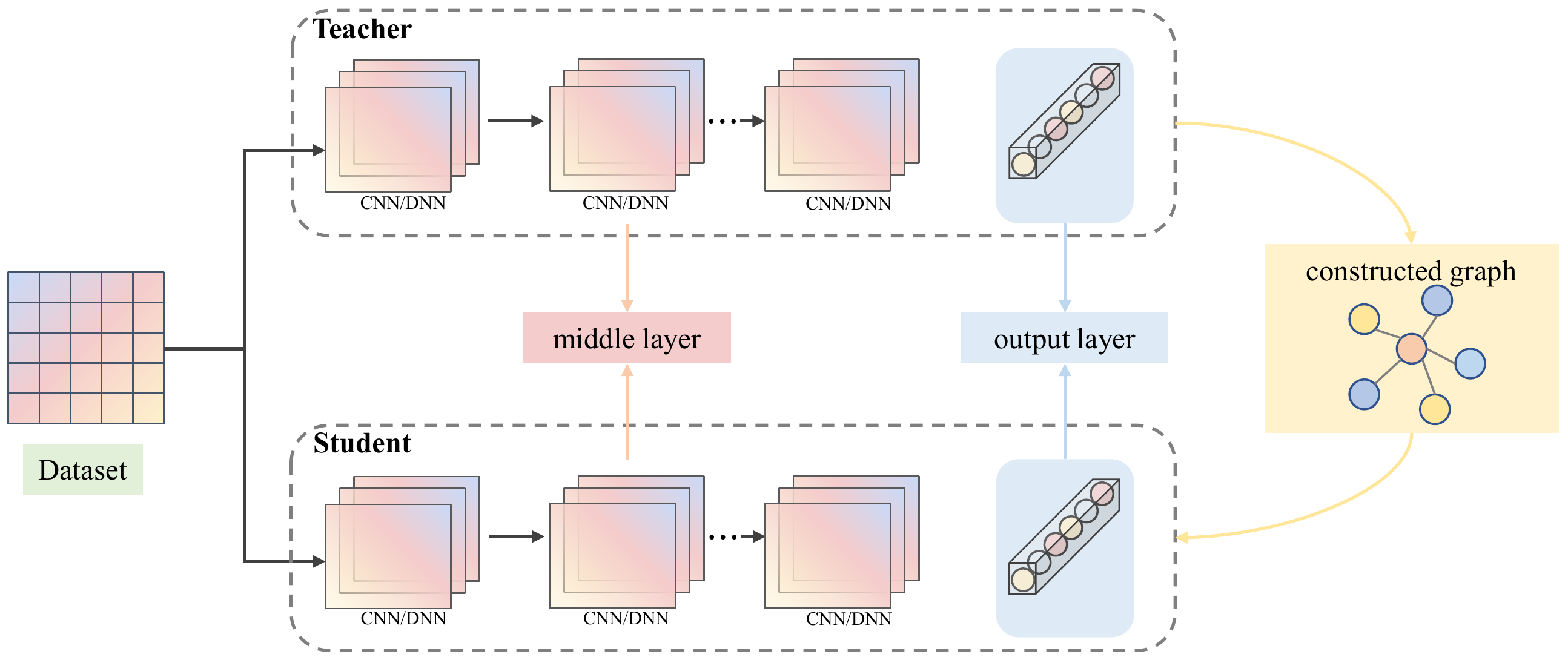}
  \caption{The framework of \zty{\zty{Graph-based Knowledge Distillation} for Deep Neural Networks} (DKD).}
  \label{fig:frame-dkd}
\end{figure}

Unfortunately, the traditional knowledge distillation method in deep learning (as shown in Eq.~\ref{eq:kd-loss}) mostly focuses on learning for individual samples. To further enhance performance, researchers proposed feature-based distillation~\cite{FitNet,AT,FT,Alp-kd} and relation-based distillation~\cite{yim2017gift,passalis2020probabilistic,DKWISL,KTG,DGCN,SPG,GCLN} methods. Among them, the most significant distillation effect is the relation-based knowledge distillation method, called the implicit constructed graph method in this paper. Subsequently, many explicitly constructed \zty{Graph-based Knowledge Distillation} methods have emerged in the intermediate convolutional layer or output layer, trying to make the student model simulate the similarity between samples in the teacher model instead of simulating the output of a single sample in the teacher model. Therefore, with the help of the implicit/explicit sample-relational constructed graph, the student model can fully mine the structured feature information between samples in the teacher network and realize the general, rich, and sufficient knowledge extracted from the teacher model to guide the student model. The framework of DKD is shown in Fig.~\ref{fig:frame-dkd}.

\textbf{Implementation:}
1) First of all, based on the sample feature representation obtained by the teacher and student model under the CNN/DNN framework, respectively, their respective sample relationship graphs (as shown in Fig.~\ref{fig:frame-dkd}) are constructed. Note that vertices of different colors represent different training samples. 2) Secondly, the similarity function is used to calculate the similarity between teacher and student network samples, respectively. 3) Finally, the distance measurement function is used to minimize the feature distribution of students and teachers to ensure that the student model can learn the correlation of multiple samples in the feature space of the teacher model.

\textbf{Formalization:}
Consequently, the final loss of the \zty{Graph-based Knowledge Distillation} for deep neural networks (DKD) is illustrated as:
\begin{equation} 
  \mathcal{L }_{D} = D_{D}\left( {\psi_{t}\left( {x_{s_{i}},x_{s_{j}}} \right),\psi_{s}\left( {x_{t_{i}},x_{t_{j}}} \right)} \right).
  \label{eq:dkd-loss}
\end{equation}
where $x_{s_{i}}$ and $x_{s_{j}}$ represent two samples i,j of the student. Similarly, $x_{t_{i}}$ and $x_{t_{j}}$ indicate two samples of the teacher. $\psi_{t}( \cdot )$ and $\psi_{s}( \cdot)$ denote the similarity function (such as cosine similarity, Jaccard similarity, etc.) between samples in the student and teacher networks, respectively. $D_G$ means to minimize the distance measurement function of the constructed graph in the student and teacher network, which can be any distance function, such as Euclidean distance, MSE, KL, etc.

\textbf{Insight and strength:}
As depicted in Fig.~\ref{fig:frame-dkd}, Most of the implicit/explicit constructed graph methods occur on the middle convolutional layer. The student model can directly extract the rich inter-sample correlation knowledge learned by the teacher model by using the constructed graph instead of just fitting the output probability distribution of an individual sample in the teacher model. The benefit of this is that the student model can capture the knowledge of spatial geometry between the input samples of the teacher model to measure the similarity between the sample features more accurately and improve the knowledge distillation learning effect of the student model. Among the knowledge distillation methods of deep neural networks, the method of knowledge distillation based on constructed graph~\cite{DKWISL,KTG,DGCN,SPG,GCLN,IEP,HKD,MHGD,IRG,DOD,HKDIFM,KDExplainer,TDD,DualDE,CAG,GKD,MorsE,BAF,LAD,GD,GCMT,GraSSNet,LSN,IntRA-KD,RKD,CC,SPKD,KCAN} has become a research hotspot at present (Related work will be classified and introduced in detail in the next Section~\ref{sec:Methodology} according to different positions of knowledge distillation). 
However, how to correctly and properly construct the auxiliary graph to model the structural knowledge of relational data is still a challenging study.

\subsection{\zty{\zty{Graph-based Knowledge Distillation} for Graph Neural Networks}}
This section begins with a simplified description of the embedding learning of GNNs. Graph, as a common data structure, is widely used to describe all kinds of relational data. A graph can be represented as $G = \left( {V, E} \right)$, where $V$ is the set of vertices, $|V| = N$ means the total number of nodes on the graph, and $E$ is the set of edges. Meanwhile, $x$ is used to represent the node feature on G, where $x \in R^{N \times D}$. $x_{i} \in R^{D}$ represents the features of the $i$-th node, and $x_{ij}$ represent the $j$-th feature of the $i$-th node. Graph representation learning mainly follows the message-passing paradigm~\cite{MPNN}: For each node $v \in V$, the node feature is updated by multi-layer aggregation of its neighborhood and node features. After $l$ iterations, the node embedding of $v$ can be obtained as follows:

\begin{equation} 
  h_{v}^{l} = \sigma\left( {h_{v}^{l - 1}~,{AGG}_{u \in N_{v}}\Phi\left( {h_{v}^{l - 1},h_{u}^{l - 1}} \right)} \right),
  \label{eq:gnn}
\end{equation}
where $u$ is the neighbor node of $v$, and $N_v$ is the set of neighborhood nodes of $v$; $AGG$ is an aggregate function such as sum, mean, or max; $\Phi$ is a message function such as MLPs (Multi-Layer Perceptrons); $\sigma$ is an update function, such as the ReLU activation function.

Although GNNs have become the current research hotspot of graph data mining and have been successfully applied in industrial fields such as medicine, recommendation, and chip design. The performance of GNNs relies heavily on a large amount of high-quality labeled data and highly complex network models. In order to obtain a GNN model with strong generalization ability, researchers focus on designing knowledge distillation algorithms on graphs, combining knowledge distillation with GNNs.

\begin{figure}[htbp]
    \centering
  \includegraphics[width=0.5\textwidth]{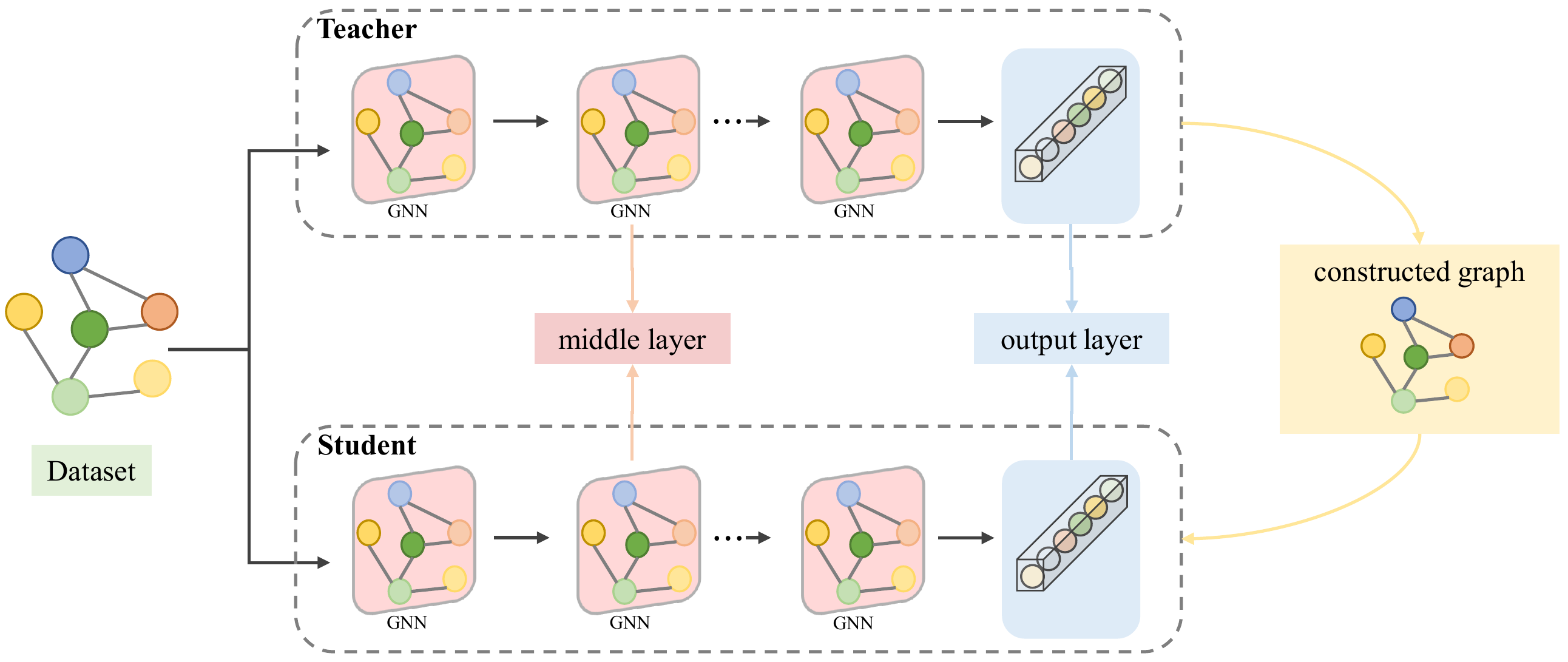}
  \caption{The framework of \zty{\zty{Graph-based Knowledge Distillation} for Graph Neural Networks} (GKD).}
  \label{fig:frame-gkd}
\end{figure}

Compared with the constructed graphs for \zty{Graph-based Knowledge Distillation} in DNNs, GNNs, as a powerful unstructured modeling tool, can directly model graph data to naturally mine the graph structure knowledge information in the teacher model and transfer it to the student model. The \zty{Graph-based Knowledge Distillation} method for graph neural networks (GKD) is similar to the method in DNNs, which also extracts knowledge from the graph convolution middle layer/output layer. To further analyze the relationship between nodes of the input graph, GNNs will dig deeper into the topology structure and node relational information of the teacher model by using the constructed graph. The framework of GKD is illustrated in Fig.~\ref{fig:frame-gkd}.

\textbf{Implementation:} 
Similarly, GKD is mainly divided into the following four steps: 1) Firstly, based on the intermediate feature representation of teacher and student models with GNNs framework, construct their respective inter-node relationship graphs (as depicted in Fig~\ref{fig:frame-gkd}), in which different colors represent different nodes (different types of heterogeneous nodes are represented in the heterogeneous graph). 2) Secondly, the similarity function is adopted to measure the correlation between the internal topological nodes of the teacher and student networks. 3) Then, the distance measurement function is employed to calculate the difference between the respective internal node embeddings of the teacher and students. 4) Finally, all the losses utilized to transfer the knowledge layer are accumulated, and the topological knowledge and node relational knowledge is migrated into the student model.

\textbf{Formalization:}
Therefore, the final loss of \zty{Graph-based Knowledge Distillation} for graph neural networks (GKD) is as follows:

\begin{equation} 
  \mathcal{L }_{G} = {\sum\limits_{l \in L}{\sum\limits_{{({x,x^{'}})} \in x^{2}}{D_{G}\left( {S\left( {x_{s}^{l},{x^{'}}_{s}^{l}} \right),S\left( {x_{t}^{l},{x^{'}}_{t}^{l}} \right)} \right)}}}.
  \label{eq:gkd-loss}
\end{equation}
where $x_{s}^{l}$ and ${x^{'}}_{s}^{l}$ indicate the two nodes in the GNN $l$-layer student network model. Similarly, $x_{t}^{l}$ and ${x^{'}}_{t}^{l}$ denote two nodes in the GNN $l$-layer teacher network model. S represents the similarity function between nodes in GNN convolutional layer/output layer, and $D_G$ means the distance metric function that minimizes the constructed graph in students and teachers, such as Huber, MSE, KL, MAE, etc.

\textbf{Insight and strength:}
Compared with the DKD method, the GKD differs most: GNN is a powerful tool for modeling graphs, which can be directly distilled in the middle/output layer and transfer the topological knowledge between graph nodes to the student model. To further explore the relationship between local nodes in the feature space, a lot of efforts have been made to construct the relational graph between nodes in the middle convolutional layer to extract the correlation knowledge between, such as LSP~\cite{LSP}, HIRE~\cite{HIRE}, etc. The successful application of knowledge distillation in GNNs has attracted widespread attention from academia and industry. A great deal of work has emerged~\cite{GFKD,RDD,GKD2,GLNN,Distill2Vec,MT-GCN,TinyGNN,GLocalKD,SCR,ROD,EGNN,LWC-KD,MustaD,EGAD,AGNN,ColdBrew,PGD,OAD,CKD,BGNN,EGSC,HSKDM,GRL,GFL,HGKT,CPF,LSP,scGCN,MetaHG,G-CRD,HIRE}. This paper summarizes it as GKD, and the related work will be divided and introduced in detail in Section~\ref{sec:Methodology} according to the different positions of knowledge distillation. However, how to fully dig graph topology and semantic information for knowledge transfer on GNNs is still challenging.

\subsection{\zty{Self-Knowledge Distillation based \zty{Graph-based Knowledge Distillation}}}
\begin{figure}[htbp]
    \centering
  \includegraphics[width=0.5\textwidth]{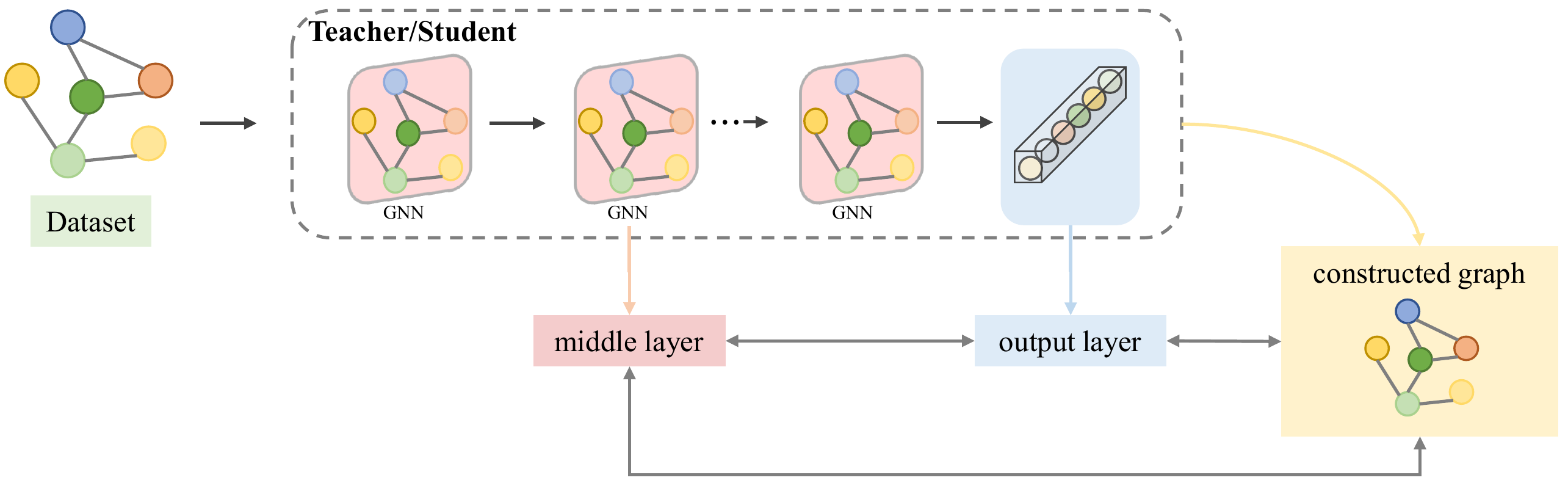}
  \caption{The framework of \zty{Self-Knowledge Distillation based \zty{Graph-based Knowledge Distillation}} (SKD).}
  \label{fig:frame-skd}
\end{figure}

\zty{Self-Knowledge Distillation} is a special case of \zty{Graph-based Knowledge Distillation} method based on T-S architecture, which refers to a special distillation method for knowledge transfer without the help of an additional teacher model. \zty{Self-Knowledge Distillation}, as the name implies, means that a single network model is both a student model and a teacher model. It usually transfers information between its own deep and shallow layers to guide its own learning without the assistance of the teacher model. Compared with the two-stage T-S distillation method, the \zty{Self-Knowledge Distillation} method is simple and efficient, which has become the first choice in the current practical landing projects. Its conceptual framework is drawn in Fig.~\ref{fig:frame-skd} (Note that this paper only summarizes the \zty{Self-Knowledge Distillation} method on the graph neural networks, excluding the \zty{Self-Knowledge Distillation} method in the deep neural networks).

\textbf{Implementation:}
The implementation steps of SKD are as follows: 1) First, based on the middle layer/output layer feature representation under the GNNs framework, construct the inter-node relational graph (as shown in Fig.~\ref{fig:frame-skd}), where different colors represent different nodes (different types of heterogeneous nodes are represented in the heterogeneous graph). 2) Then, the similarity function is leveraged to measure the similarity between the internal topological node representations of the shallow and deep layers of GNNs in the feature space. 3) Finally, the difference between shallow and deep networks is calculated by using the distance metric function, and more diverse knowledge can be learned through multiple iterative calculations.

\textbf{Formalization:}
Accordingly, the final loss of \zty{Self-Knowledge Distillation} based \zty{Graph-based Knowledge Distillation} (SKD) is denoted as:

\begin{equation} 
 \mathcal{L }_{Self} = {\sum_{l = 1,\ldots,l - 1}{D_{Self}\left( {G_{s}^{l}(X),G_{t}^{l + 1}(X)} \right)}}.
  \label{eq:skd-loss}
\end{equation}
where $X$ represents the whole graph node representation, $G_{s}^{l}(X)$ denotes the inter-node relational knowledge constructor in the GNN shallow layer $l$. Likewise, $G_{t}^{l + 1}(X)$ represents the inter-node relational knowledge constructor in the GNN deep layer $l+1$. $D_{self}$ represents the distance metric function of the shallow layer and deep layer constructed graph, such as InfoCE, KL, MSE, etc.

\textbf{Insight and strength:}
Compared with the traditional two-stage T-S distillation mode, the \zty{Self-Knowledge Distillation} learning mode can greatly save the model training time, greatly improve the training efficiency, and achieve model performance improvement without the teacher's guidance. However, \zty{Self-Knowledge Distillation} also has certain shortcomings: (1) It lacks abundant external knowledge, which may help to improve the performance of student models if external knowledge can be explicitly introduced, such as combined with knowledge graphs. (2) There is no conclusion on the advantages and disadvantages of the traditional two-stage T-S distillation mode and \zty{Self-Knowledge Distillation} mode, and there is a lack of comparative analysis of them in the same experimental environment and task, which is worthy of further study. (3) The interpretability analysis of \zty{Self-Knowledge Distillation} needs further study. At present, the \zty{Self-Knowledge Distillation} on CNN/DNN focuses on the deep-layer to shallow distillation, while on GNNs, it is the opposite and lacks theoretical support. Therefore, the versatility and flexibility of SKD need to be further explored.

\section{Methodology}\label{sec:Methodology}
\subsection{\zty{\zty{Graph-based Knowledge Distillation} for Deep Neural Networks}}
The core of knowledge distillation lies in the extraction of knowledge, while knowledge exists in different positions in the model. Therefore, in accordance with the position of knowledge distillation, Graph-based Knowledge Distillation for Deep Neural Networks (DKD) is divided into the output layer, middle layer, and constructed graph knowledge. This section mainly introduces these three types of knowledge transfer forms, and the relevant graph distillation methods introduced below are all based on this. See TABLE~\ref{tab:dkd-taxonom} for details. (Note that only the most prominent forms of knowledge distillation are highlighted in the description of various \zty{Graph-based Knowledge Distillation} methods.)

\textbf{Similarities:}
For the DKD methods based on the output layer, middle layer, and constructed graph, each type of \zty{Graph-based Knowledge Distillation} algorithm is based on knowledge extraction at the same location. They all follow the DKD model framework of Fig.~\ref{fig:frame-dkd} and the graph distillation loss  $L_D$ paradigm of Eq.~\ref{eq:dkd-loss}.

\textbf{Differences:}
For each type of DKD, their differences are reflected in many aspects, such as specific implementation, distance metric function, downstream tasks, applications, etc.  For details, see TABLE~\ref{tab:dkd-taxonom} below.  For example, in DKD methods based on output layer knowledge: DKWISL~\cite{DKWISL} applies KD to relation extraction in NLP by using the KL distance metric.  KTG~\cite{KTG} is used for image recognition applications of collaborative learning by using KL to measure the distribution difference between teachers and students, while GCLN~\cite{GCLN} utilizes $L_2$ to apply KD to the visual robot localization scenarios for image semantic segmentation.  Among the methods of DKD based on middle-layer knowledge: IEP~\cite{IEP} combines KL and L1 to apply knowledge to transfer learning and image classification on multi-task learning.  HKD~\cite{HKD} leverages InfoCE to introduce knowledge distillation technology in visual dialogue tasks of graph inference, and IRG~\cite{IRG} proposes Hit to apply knowledge distillation to image recognition scenarios.  Similarly, DKD methods based on constructed graph knowledge are similar: CAG~\cite{CAG} presents the constructed \zty{Graph-based Knowledge Distillation} technique to enhance the visual dialogue performance of the student model in the downstream graph reasoning task.  GKD~\cite{GKD} explores Frobenius to minimize the distribution difference between teachers and students and compresses the student model.  MorsE~\cite{MorsE} employs $L_2$ to transfer meta-knowledge to improve the student model in the link prediction and question-answering system tasks.

\begin{table*}[htbp]
    \caption{A taxonomy of \zty{\zty{Graph-based Knowledge Distillation} for Deep Neural Networks (DKD)}.}
    \centering
    \resizebox{1.0\linewidth}{!}{
    \begin{tabular}{c|ccc|c|c|c}
    \toprule
        \multirow{2}{*}{Method} & \multicolumn{3}{c|}{Distillation Location}                                                      & \multirow{2}{*}{Distance Measurement} & \multirow{2}{*}{Task}          & \multirow{2}{*}{Application}                   \\ \cline{2-4}
                        & \multicolumn{1}{c|}{Output Layer} & \multicolumn{1}{c|}{Intermediate Layer} & Constructed Graph &                                       &                                &                                                \\ \hline
        IEP~\cite{IEP}             & \multicolumn{1}{c|}{}             & \multicolumn{1}{c|}{\checkmark}                  &                   & KL, L1                                & Multi-task learning            & Transfer learning, image classification        \\ \hline
        HKD~\cite{HKD}             & \multicolumn{1}{c|}{}             & \multicolumn{1}{c|}{\checkmark}                  &                   & InfoCE                                & Knowledge distillation         & Image classification, knowledge transfer       \\ \hline
        CAG~\cite{CAG}             & \multicolumn{1}{c|}{}             & \multicolumn{1}{c|}{}                   & \checkmark                 & KL                                    & Graph inference                & Visual dialogue                                \\ \hline
        DKWISL~\cite{DKWISL}         & \multicolumn{1}{c|}{\checkmark}            & \multicolumn{1}{c|}{}                   &                   & KL                                    & Natural language processing    & Relation extraction                            \\ \hline
        KTG~\cite{KTG}            & \multicolumn{1}{c|}{\checkmark}            & \multicolumn{1}{c|}{}                   &                   & KL                                    & Collaborative learning         & Image recognition                              \\ \hline
        MHGD~\cite{MHGD}           & \multicolumn{1}{c|}{}             & \multicolumn{1}{c|}{\checkmark}                  &                   & KL                                    & Multi-task learning            & Image recognition                              \\ \hline
        IRG~\cite{IRG}            & \multicolumn{1}{c|}{\checkmark}            & \multicolumn{1}{c|}{\checkmark}                  &                   & Hit                                   & Knowledge distillation         & Image recognition                              \\ \hline
        DGCN~\cite{DGCN}           & \multicolumn{1}{c|}{\checkmark}            & \multicolumn{1}{c|}{}                   &                   & KL                                    & Collaborative filtering        & Item recommendations                        \\ \hline
        GKD~\cite{GKD}            & \multicolumn{1}{c|}{\checkmark}            & \multicolumn{1}{c|}{\checkmark}                  & \checkmark                 & Frobenius                             & Model compression              & Image classification                           \\ \hline
        SPG~\cite{SPG}           & \multicolumn{1}{c|}{\checkmark}            & \multicolumn{1}{c|}{}                   &                   & KL                                    & Natural language processing    & Video captioning                               \\ \hline
        MorsE~\cite{MorsE}           & \multicolumn{1}{c|}{}             & \multicolumn{1}{c|}{}                   & \checkmark                 & $L_2$                                    & Meta-knowledge transfer        & Link prediction, question answering system     \\ \hline
        GCLN~\cite{GCLN}           & \multicolumn{1}{c|}{\checkmark}            & \multicolumn{1}{c|}{}                   &                   & $L_2$                                    & Image semantic segmentation    & Vision robot self-positioning                  \\ \hline
        DOD~\cite{DOD}           & \multicolumn{1}{c|}{\checkmark}            & \multicolumn{1}{c|}{\checkmark}                  &                   & KL                                    & Object detection               & Object Detectors                               \\ \hline
        BAF~\cite{BAF}            & \multicolumn{1}{c|}{}             & \multicolumn{1}{c|}{}                   & \checkmark                 & EMD                                   & Model compression              & Video classification                           \\ \hline
        LAD~\cite{LAD}             & \multicolumn{1}{c|}{}             & \multicolumn{1}{c|}{}                   & \checkmark                 & BELU                                  & Natural language processing    & Machine translation                            \\ \hline
        GD~\cite{GD}            & \multicolumn{1}{c|}{}             & \multicolumn{1}{c|}{}                   & \checkmark                 & Cosine                                & Multimodal video               & Motion detection, action classification        \\ \hline
        GCMT~\cite{GCMT}            & \multicolumn{1}{c|}{\checkmark}            & \multicolumn{1}{c|}{\checkmark}                  & \checkmark                 & CE                                    & Unsupervised domain adaptation & Person re-identification                       \\ \hline
        GraSSNet~\cite{GraSSNet}       & \multicolumn{1}{c|}{}             & \multicolumn{1}{c|}{}                   & \checkmark                 & MSE                                   & Knowledge transfer             & Saliency prediction                            \\ \hline
        LSN~\cite{LSN}             & \multicolumn{1}{c|}{\checkmark}            & \multicolumn{1}{c|}{\checkmark}                  & \checkmark                 & KL, MSE                               & Model compression              & Node classification                            \\ \hline
        IntRA-KD~\cite{IntRA-KD}        & \multicolumn{1}{c|}{}             & \multicolumn{1}{c|}{\checkmark}                  & \checkmark                 & MSE                                   & Model compression              & Road marking segmentation                      \\ \hline
        RKD~\cite{RKD}            & \multicolumn{1}{c|}{\checkmark}            & \multicolumn{1}{c|}{}                   & \checkmark                 & Euclidean,Huber                       & Knowledge distillation         & Image classification, few-Shot Learning        \\ \hline
        CC~\cite{CC}             & \multicolumn{1}{c|}{\checkmark}            & \multicolumn{1}{c|}{\checkmark}                  & \checkmark                 & KL, MSE                               & Knowledge distillation         & Image classification, person re-identification \\ \hline
        SPKD~\cite{SPKD}            & \multicolumn{1}{c|}{}             & \multicolumn{1}{c|}{}                   & \checkmark                 & Frobenius                             & Knowledge distillation         & Image classification, transfer learning        \\ \hline
        HKDIFM~\cite{HKDIFM}         & \multicolumn{1}{c|}{}             & \multicolumn{1}{c|}{\checkmark}                  &                   & KL                                    & Knowledge distillation         & Image classification                           \\ \hline
        KDExplainer~\cite{KDExplainer}     & \multicolumn{1}{c|}{\checkmark}            & \multicolumn{1}{c|}{\checkmark}                  &                   & CE, KL                                & Interpretability               & Image classification                           \\ \hline
        TDD~\cite{TDD}            & \multicolumn{1}{c|}{\checkmark}            & \multicolumn{1}{c|}{\checkmark}                  &                   & CE, KL                                & Interpretability               & Image classification                           \\ \hline
        DualDE~\cite{DualDE}          & \multicolumn{1}{c|}{}             & \multicolumn{1}{c|}{\checkmark}                  &                   & JSD                                   & Knowledge distillation         & Node classification, link prediction           \\ \hline
        KCAN~\cite{KCAN}            & \multicolumn{1}{c|}{}             & \multicolumn{1}{c|}{}                   & \checkmark                 & BPR                                   & Knowledge graph                & Top-K Recommendation, TR Prediction            \\ 
        \bottomrule
    \end{tabular}
    }
    \label{tab:dkd-taxonom}
\end{table*}

\subsubsection{Output Layer Knowledge}
Knowledge distillation based on the output layer is the label supervision information contained in the predicted output of the last layer on the network model, which is currently the most popular form of knowledge distillation in DNNs. Since KD~\cite{KD} method was proposed, such methods have received much attention from scholars, and a lot of excellent work has been derived subsequently. This section focuses on the output layer knowledge of CNN/DNN with the help of a constructed graph. That is, the DKD methods consider the relationship between samples.

The earliest DKD method of output layer can be traced back to KTG, a method of knowledge transfer control based on a constructed graph proposed by Minami \etal~\cite{KTG}, which represents different patterns of knowledge transfer through a unified view of knowledge transfer. It also introduces four types of gate functions to control backpropagation during network training to explore different knowledge transfer combination forms. In the same year, Wang \etal~\cite{DGCN} distill the ranking information from the GCN model into the binary model with the help of GNNs to fully mine rich connection information between users and items in the commodity data and successfully realized the performance improvement of the online recommendation model and the acceleration of implicit feedback recommendation. Meanwhile, Zhang \etal~\cite{DKWISL} also present to combine the soft label of the output layer with the GCN model, successfully applying the knowledge of the output layer to natural language processing. Specifically, DKWISL first obtains type-restricted soft rules from the entire corpus, and then the teacher model combines the designed soft rules with GCN to get the final soft label for each instance.

Additionally, some related work has been proposed. Pan \etal~\cite{SPG} argue that the previous video description model has not clearly modeled the interaction between objects and propose a novel spatiotemporal graph network that explicitly uses spatiotemporal object interaction and introduce a knowledge distillation mechanism SPG with object perception to regularize global scene features by using local object information, which solves the problem of noise feature learning in a spatiotemporal graph model. Koji \etal~\cite{GCLN} apply output layer knowledge to robot self-localization by designing a Teacher-Student knowledge transfer scheme GCLN based on rank matching, where the reciprocal rank vector output of the existing teacher self-localization model was transferred to the student model as dark knowledge.

\subsubsection{Middle Layer Knowledge}
Considering the monolithic nature of the knowledge distillation method of the output layer, to further explore the rich knowledge contained in the teacher network, many researchers have begun to study how to transfer the feature knowledge in the intermediate convolutional layer to the student network to obtain a high-quality feature representation.   FitNets~\cite{FitNet} is the first method to use mid-layer feature distillation, aiming to use the middle-layer output of the teacher model feature extractor as hints to distill the knowledge of deeper and narrower student models.   It is different from existing intermediate-layer feature distillation methods.   This section focuses on summarizing the distillation method on DNNs based on the knowledge of middle-layer feature relationships.   Among such methods, the most representative work is IRG proposed by Liu \etal~\cite{IRG} in the 2019 CVPR.   Unlike considering only instance features knowledge, IRG also introduces two additional kinds of knowledge: instance relationships and feature space transformation.

The success of IRG has spawned a lot of related work. For example, Lee \etal~\cite{MHGD} utilize the MHGD method that uses a multi-head attention mechanism to extract knowledge in teachers' embedding process as a graph, to empower students with relational inductive bias through multi-tasking learning. Passalis \etal~\cite{HKDIFM} observe that the student model simulates the information flow of the teacher model at the key learning stage to ensure the necessary connections between the layers of the network, aiming to achieve effective knowledge transfer. Lee \etal~\cite{IEP} explore an explainable embedding process (IEP) knowledge generation method based on principal component analysis, extract the knowledge by MPNN~\cite{MPNN}, and confirm the interpretability of IEP for embedding process knowledge through visualization. Zhou \etal~\cite{HKD} employ to extract the overall knowledge from the teacher network based on the attribute graph between the instances, realize the fusion of individual knowledge and relational knowledge, and retain the correlation between the two kinds of knowledge so that the student model can obtain sufficient knowledge during training. Xue \etal~\cite{KDExplainer} propose KDExplainer to elucidate the working mechanism of soft targets in the KD process and find that KD can implicitly adjust knowledge conflicts between subtasks, which is more effective than label smoothing.    Song \etal~\cite{TDD} present a new tree-like decision distillation (TDD), which analyzes the teacher's decision-making process through layer-wise mode and imposes the same decision-making constraints on the student model to promote students to master the same problem-solution. Zhu \etal~\cite{DualDE} explore a novel distillation method called DualDE, which distills teachers' and students' triplet output scores and intermediate layer embedded structural knowledge to each other and introduces a soft label evaluation mechanism to evaluate the soft label quality provided by teachers/students to achieve dual optimization of students and teachers. Chen \etal~\cite{DOD} design a structured case graph based on the relationship between each region of interest (RoI), instance,  and at the same time, used the similarities between instance features and features to transfer knowledge in a structured way, ensuring that students' models could capture global topology structure knowledge and soft label knowledge.

The distillation method based on middle-layer knowledge can not only distill the knowledge of an individual sample but also distill the knowledge of the relationship between the spatial samples in the student network, which has become one of the essential methods for distilling graph knowledge in the current DNNs.

\subsubsection{Constructed Graph Knowledge}
To better model the supervisory information of the relationship between samples in the teacher network, the constructed graph knowledge method is proposed. The knowledge distillation method based on the constructed graph is an extension of the knowledge method of the middle layer. It aims to dig the high-order correlation knowledge among sample features in the teacher network by constructing an explicit auxiliary graph structure module. RKD~\cite{RKD}, a representative work, was proposed by Park \etal in 2019, which experimentally proves that extracting the relationship structure information between samples is better than extracting the feature information of a single sample.

Since RKD is proposed, the knowledge distillation method based on the constructed graph has become a research hotspot of DKD. The related DKD works based on the constructed graph are applied. For example, Zhang \etal~\cite{BAF} propose to construct two auxiliary graphs using logits and intermediate features to transfer the knowledge of multiple self-supervised teachers to students and achieve the goal of model compression. Chen \etal~\cite{LSN} view the Teacher-Student distillation paradigm from a new perspective of feature embedding and maintain the relationship knowledge between samples in high-dimensional space in teacher networks by introducing local location retention loss LSN with the help of constructed graphs.   Peng \etal~\cite{CC} believe that the correlation between instances is also valuable knowledge to enhance students' performance. Therefore, a distillation framework CC is proposed, which not only considers output layer knowledge and intermediate knowledge but also uses graph-based knowledge (association information between instances) when transferring knowledge. Tung \etal~\cite{SPKD} observe that similar semantic inputs often lead to similar activation patterns and then introduce SPKD to guidance in the training of student networks, making students not need to imitate the teacher's representation space but only retain the similarity in their own representation space. Lassance \etal~\cite{GKD} extend RKD to capture geometric information about potential spaces with the help of constructed graphs, thereby transferring knowledge from teacher architecture to student networks.

Besides, many scholars have explored the application of KD in other fields. He \etal~\cite{LAD} introduce the concept of the linguistic graph and propose the graph distillation algorithm LAD to improve the accuracy of machine translation. Luo \etal~\cite{GD} present a cross-modal dynamic distillation method GD, combining graph distillation and domain transfer techniques to realize knowledge transfer between multimodal. Liu \etal~\cite{GCMT} adopt the GCMT method based on graph consistency for adaptive pedestrian re-identification in unsupervised domains. Zhang \etal~\cite{GraSSNet} propose GraSSNet to encode the semantic relations learned from external knowledge with the help of constructed auxiliary graphs to achieve significance prediction. Tu \etal~\cite{KCAN} study how to introduce external knowledge graph structure information into the recommendation network and distill the knowledge graph by spreading personalized information on the sampling subgraph through local conditional attention. Chen \etal~\cite{MorsE} perform a new task meta-knowledge transfer method MorsE for knowledge graph tasks (KGs), achieving excellent performance in link prediction and question-answering systems. Guo \etal~\cite{CAG} utilize a fine-grained context-aware graph (CAG) distillation scheme for visual dialogue to solve the noise problem in visual dialogue tasks. Hou \etal~\cite{IntRA-KD} successfully apply the distillation of constructed graph knowledge to road marking segmentation scenarios.

Although the DKD methods show great potential and are successfully applied to a variety of downstream tasks, these methods focus on CNN/DNN and cannot directly apply data with graphs. Recently, a large number of researchers have tried to apply KD to GNNs with impressive results, which will be elaborated next section.

\subsection{\zty{\zty{Graph-based Knowledge Distillation} for Graph Neural Networks}}
Similarly, for the knowledge distillation method in GNNs, this section still classifies them as output layer knowledge-based, middle layer knowledge-based, and constructed graph knowledge-based according to the location of knowledge distillation. This section delineates Graph-based Knowledge Distillation for Graph Neural Networks (GKD) methods in detail, as shown in TABLE~\ref{tab:gkd-taxonomy} below.
(Note that only the most prominent forms of knowledge distillation are highlighted in the description of various \zty{Graph-based Knowledge Distillation} methods.)

\textbf{Similarities:}
For the GKD methods based on output layer, middle layer, and constructed graph, each type of \zty{Graph-based Knowledge Distillation} algorithm is based on the extraction of knowledge at the same location. For the DKD methods based on output layer, middle layer, and constructed graph, each type of \zty{Graph-based Knowledge Distillation} algorithm is based on knowledge extraction at the same location. They all follow the DKD model framework of Fig.~\ref{fig:frame-gkd} and the graph distillation loss  $L_G$ paradigm of Eq.~\ref{eq:gkd-loss}.

\textbf{Differences:}

TABLE~\ref{tab:gkd-taxonomy} shows the differences between various types of GKD. Output layer-based methods using KL divergence are GFKD~\cite{GFKD}, GLNN~\cite{GLNN}, Distill2Vec~\cite{Distill2Vec}, and MT-GCN~\cite{MT-GCN}; middle layer-based methods using KL divergence are MustaD~\cite{MustaD}, OAD~\cite{OAD}, BGNN~\cite{BGNN}, and HSKDM~\cite{HSKDM}; constructed graph-based methods using KL divergence are CPF~\cite{CPF}, LSP~\cite{LSP}, MetaHG~\cite{MetaHG}, and HIRE~\cite{HIRE}. Similarly, output layer-based methods using MSE are RDD~\cite{RDD} and SCR~\cite{SCR}; middle layer-based methods using MSE are LWC-KD~\cite{LWC-KD}, AGNN~\cite{AGNN}, Cold Brew~\cite{ColdBrew}, and EGSC~\cite{EGSC}. Finally, HIRE~\cite{HIRE} uses the constructed graph knowledge of a relational metric types to measure the differences between teacher and student models.

\begin{table*}[htbp]
    \caption{A taxonomy of \zty{\zty{Graph-based Knowledge Distillation} for Graph Neural Networks (GKD)}.}
    \centering
    \resizebox{1.0\linewidth}{!}{
    \begin{tabular}{c|ccc|c|c|c}
    \toprule
        \multirow{2}{*}{Method} & \multicolumn{3}{c|}{Distillation Location}                                                      & \multirow{2}{*}{Distance Measurement} & \multirow{2}{*}{Task}           & \multirow{2}{*}{Application}                      \\ \cline{2-4}
                        & \multicolumn{1}{c|}{Output Layer} & \multicolumn{1}{c|}{Intermediate Layer} & Constructed Graph &                                       &                                 &                                                   \\ \hline
        GFKD~\cite{GFKD}           & \multicolumn{1}{c|}{\checkmark}            & \multicolumn{1}{c|}{}                   &                   & KL                                    & Data-free distillation          & Zero-Shot learning                                \\ \hline
        LWC-KD~\cite{LWC-KD}        & \multicolumn{1}{c|}{}             & \multicolumn{1}{c|}{\checkmark}                  &                   & MSE, $L_2$                               & Incremental learning            & Recommender system                                \\ \hline
        MustaD~\cite{MustaD}        & \multicolumn{1}{c|}{\checkmark}            & \multicolumn{1}{c|}{\checkmark}                  &                   & KL                                    & Model compression               & Node classification                               \\ \hline
        RDD~\cite{RDD}             & \multicolumn{1}{c|}{\checkmark}            & \multicolumn{1}{c|}{}                   &                   & MSE                                   & Semi-supervised learning        & Node classification                               \\ \hline
        EGAD~\cite{EGAD}           & \multicolumn{1}{c|}{}             & \multicolumn{1}{c|}{\checkmark}                  &                   & RMSE, MAE                             & Semi-supervised learning        & Live video streaming events                       \\ \hline
        GRL~\cite{GRL}            & \multicolumn{1}{c|}{}             & \multicolumn{1}{c|}{}                   & \checkmark                 & MAE                                   & Multi-task learning             & Graph-level prediction                            \\ \hline
        GFL~\cite{GFL}           & \multicolumn{1}{c|}{}             & \multicolumn{1}{c|}{}                   & \checkmark                 & Frobenius                             & Few-Shot learning               & Node classification                               \\ \hline
        HGKT~\cite{HGKT}           & \multicolumn{1}{c|}{}             & \multicolumn{1}{c|}{}                   & \checkmark                 & Wasserstein                           & Zero-Shot learning              & Node classification                               \\ \hline
        AGNN~\cite{AGNN}          & \multicolumn{1}{c|}{\checkmark}            & \multicolumn{1}{c|}{\checkmark}                  &                   & MSE                                   & Model compression               & Node classification, point cloud classification                                         \\ \hline
        CPF~\cite{CPF}           & \multicolumn{1}{c|}{\checkmark}            & \multicolumn{1}{c|}{\checkmark}                  & \checkmark                 & $L_2$, KL                                & Knowledge distillation          & Node classification                               \\ \hline
        LSP~\cite{LSP}           & \multicolumn{1}{c|}{}             & \multicolumn{1}{c|}{\checkmark}                  & \checkmark                 & KL                                    & Model compression               & Node classification, point cloud classification   \\ \hline
        GKD~\cite{GKD2}            & \multicolumn{1}{c|}{\checkmark}            & \multicolumn{1}{c|}{}                   &                   & CE                                    & Graph inference                 & Disease diagnosis and prediction                  \\ \hline
        scGCN~\cite{scGCN}         & \multicolumn{1}{c|}{}            & \multicolumn{1}{c|}{}                   &  \checkmark                 & CE                                    & Single cell omics               & Cell identification, cross-species classification \\ \hline
        MetaHG~\cite{MetaHG}        & \multicolumn{1}{c|}{}            & \multicolumn{1}{c|}{}                   &   \checkmark                & KL                                    & Illegal drug trafficker         & Classification                                    \\ \hline
        Cold Brew~\cite{ColdBrew}      & \multicolumn{1}{c|}{\checkmark}            & \multicolumn{1}{c|}{\checkmark}                  &                   & MSE, CE                               & Cold start                      & Recommender system                                \\ \hline
        PGD~\cite{PGD}           & \multicolumn{1}{c|}{}             & \multicolumn{1}{c|}{\checkmark}                  &                   & MSE                                   & Cold start                      & Recommender system                                \\ \hline
        GLNN~\cite{GLNN}          & \multicolumn{1}{c|}{\checkmark}            & \multicolumn{1}{c|}{}                   &                   & KL                                    & Offline knowledge distillation  & Node classification                               \\ \hline
        Distill2Vec~\cite{Distill2Vec}     & \multicolumn{1}{c|}{\checkmark}            & \multicolumn{1}{c|}{}                   &                   & KL                                    & Model compression               & Link prediction                                   \\ \hline
        MT-GCN~\cite{MT-GCN}         & \multicolumn{1}{c|}{\checkmark}            & \multicolumn{1}{c|}{}                   &                   & KL                                    & Semi-supervised learning        & Node classification                               \\ \hline
        TinyGNN~\cite{TinyGNN}        & \multicolumn{1}{c|}{\checkmark}            & \multicolumn{1}{c|}{}                   &                   & CE                                    & Model compression               & Node classification                               \\ \hline
        GLocalKD~\cite{GLocalKD}        & \multicolumn{1}{c|}{\checkmark}            & \multicolumn{1}{c|}{}                   &                   & KL                                    & Anomaly detection               & Anomaly detection                                 \\ \hline
        OAD~\cite{OAD}           & \multicolumn{1}{c|}{\checkmark}            & \multicolumn{1}{c|}{\checkmark}                  &                   & CE, KL                                & Online adversarial distillation & Node classification                               \\ \hline
        SCR~\cite{SCR}            & \multicolumn{1}{c|}{\checkmark}            & \multicolumn{1}{c|}{}                   &                   & MSE                                   & Model training                  & Node classification                               \\ \hline
        ROD~\cite{ROD}            & \multicolumn{1}{c|}{\checkmark}            & \multicolumn{1}{c|}{}                   &                   & KL                                    & Model compression               & Node classification, clustering, link prediction  \\ \hline
        EGNN~\cite{EGNN}          & \multicolumn{1}{c|}{\checkmark}            & \multicolumn{1}{c|}{}                   &                   & KL                                    & Model interpretability          & Node classification                               \\ \hline
        CKD~\cite{CKD}          & \multicolumn{1}{c|}{}             & \multicolumn{1}{c|}{\checkmark}                  &                   & JSD                                   & Knowledge distillation          & Node classification, link prediction              \\ \hline
        G-CRD~\cite{G-CRD}         & \multicolumn{1}{c|}{\checkmark}            & \multicolumn{1}{c|}{\checkmark}                  & \checkmark                 & InfoCE                                & Model compression               & Classification, similarity measures               \\ \hline
        BGNN~\cite{BGNN}          & \multicolumn{1}{c|}{}             & \multicolumn{1}{c|}{\checkmark}                  &                   & KL                                    & Model compression               & Image classification                              \\ \hline
        EGSC~\cite{EGSC}          & \multicolumn{1}{c|}{}             & \multicolumn{1}{c|}{\checkmark}                  &                   & MSE, Huber                            & Model compression               & Anomaly detection, graph similarity calculation   \\ \hline
        HSKDM~\cite{HSKDM}         & \multicolumn{1}{c|}{}             & \multicolumn{1}{c|}{\checkmark}                  &                   & KL, Triplet                           & Knowledge distillation          & Node classification                               \\ \hline
        HIRE~\cite{HIRE}          & \multicolumn{1}{c|}{\checkmark}            & \multicolumn{1}{c|}{\checkmark}                  & \checkmark                 & KL, MSE                               & Knowledge distillation          & Node classification, clustering, visualization    \\ 
        \bottomrule
    \end{tabular}
    }
    \label{tab:gkd-taxonomy}
\end{table*}

\subsubsection{Output Layer Knowledge}
Inspired by the graph distillation technique in deep learning, researchers related to GNNs have successfully applied KD and GNNs technologies to various application scenarios of graph data mining, including recommendation learning, anomaly detection, and cell recognition. This section focuses on output layer knowledge-based distillation methods of GKD.

The application of KD in GNNs has attracted attention in recent years, and the output layer distillation work can be traced back to the proposal of TinyGNN~\cite{TinyGNN} in 2020. Specifically, Yan \etal propose a peer-aware module (PAM) and a neighbor distillation strategy (NDS) to explicitly and implicitly model the local node information structure, respectively, thus effectively characterizing local structures and learning better node representation.

Subsequently, researchers propose a large number of related works based on output layer knowledge.   For example, Zhang \etal~\cite{RDD} design a reliable data distillation (RDD) method to optimize traditional KD and enhance model representation capabilities by defining node reliability and edge reliability in the graph to better utilize high-quality data. Antaris \etal~\cite{Distill2Vec} develop a distillation loss function based on Kullback-Leibler divergence, transferring the acquired knowledge from a teacher model trained offline to a small student model trained online while also employing a self-attentional mechanism to capture the graph evolution in the learned node embedding. Deng \etal~\cite{GFKD} adopt a GFKD framework for knowledge distillation from GNN without graph data in considering consideration of data unavailability caused by data privacy and other issues. To achieve this goal, GFKD utilizes a structural learning strategy to model the graph topology of the multivariate Bernoulli distribution and then introduces gradient estimation to optimize it.   Ghorbani \etal~\cite{GKD} leverage the label propagation algorithm to inject all graph-related knowledge into the pseudo-labels generated by the teacher model and then use the pseudo-labels to train the student network. 
Zhan \etal~\cite{MT-GCN} present a simple and effective graph semi-supervised learning-based distillation method MT-GCN, which expands the label set with high-confidence predictions as pseudo-labels to select more samples to update the GCN model. Zhang \etal~\cite{SCR} study the role of consistency regularization in the training of semi-supervised graph neural networks through knowledge distillation and propose that model training could be guided by calculating the consistency loss between the student and teacher models.     To solve the problems of edge sparsity and label sparsity, Zhang \etal~\cite{ROD} first observe the online distillation method ROD to dynamically train a powerful teacher/student model by integrating multi-scale perceptual graph knowledge.     Ma \etal~\cite{GLocalKD} explore a novel deep anomaly detection method, GLocalKD, which learns the rich global and local graph regularization pattern information of graphs by jointly random distillation of graphs and nodes.   Based on the idea of knowledge distillation,  Li \etal~\cite{EGNN} perform an interpretable shallow graph neural network model EGNN for graph representation. A new distillation framework, GLNN, is proposed by Zhang \etal~\cite{GLNN}, greatly reducing the inference time of node classification by extracting the knowledge of logits in the teacher GNN into the student MLP model.

Although the output layer knowledge shows great advantages in GNNs, such methods are only applied to the KD framework proposed by Hinton \etal, and only use the supervised information of probability distribution on the output layer, failing to explore the knowledge in the teacher's GNN fully.

\subsubsection{Middle Layer Knowledge}
In view of the limited effect of output layer knowledge on the information extraction of teacher models in GNNs, the introduction of middle layer knowledge can further enrich the representation of knowledge and effectively improve the performance of downstream tasks of student models. This section concludes the middle layer knowledge of GKD, which is mainly divided into two categories according to downstream tasks.

One category is used in common graph mining tasks, such as node classification.     For example, Jing \etal~\cite{AGNN} exert a many-to-one teacher-student distillation framework AGNN, where multiple teacher networks are used to jointly train a student network so that students can learn from teachers with different characteristic dimensions.     Kim \etal~\cite{MustaD} integrate the aggregate information of the convolutional layer of the middle layer of the teacher network and the soft label knowledge of the output layer and extract it into the student network.     Wang \etal~\cite{OAD} employ the first online adversarial distillation method OAD for GNNs to effectively capture the structural changes of GNNs. Wang \etal~\cite{CKD} first attempt to use the collaborative knowledge distillation method to model the correlation between meta-path embeddings in heterogeneous information networks, which can effectively maintain global similarity and distill local knowledge in the final embedding process.   To improve the classification performance of GNNs in category-unbalanced graph data, Huang \etal~\cite{HSKDM} adopt a hard sample-based knowledge distillation method named HSKDM, which extracts the knowledge of the middle layer and output layer of the model into the student network by jointly training multiple GNN models.

The other category is the middle layer knowledge of GKD for other graph analysis tasks such as recommendation systems~\cite{ColdBrew,LWC-KD,PGD}.     Zheng \etal~\cite{ColdBrew} develop knowledge distillation technology to embed teacher nodes on low-dimensional manifolds by using graph structures and require students to learn the mapping from node features to the manifold, generalizing GNN to cold start research problems in recommender systems.     In the same year, Wang \etal~\cite{LWC-KD} combine contrast learning and distillation learning and propose a layer-by-layer comparative distillation framework named LWC-KD for the recommended scenario of incremental learning.    Wang \etal~\cite{PGD} utilize a new privileged graph distillation model (PGD) by using the advantages of graph learning and knowledge distillation in privileged information modeling to improve the performance of GNNs in cold start problems. Additionally, the middle layer knowledge of GKD is also used for other tasks.    For example, Anantis \etal~\cite{EGAD} first utilize EGAD for dynamic graph representation learning, introducing a weighted self-attention mechanism between continuous dynamic graph convolutional networks to capture the evolution of real-time video stream event graphs.    Bahri \etal~\cite{BGNN} introduce a binary graph neural network named BGNN based on XNOR-Net++ and knowledge distillation and also explore the influence on the image classification performance of binary graph neural networks by studying various strategies and design decisions.    Qin \etal~\cite{EGSC} design a novel multi-level GNN feature fusion model based on common attention to solving slow graph similarity learning,  which leverages the knowledge distillation method to extract knowledge distillation from the model to the student model.

In a nutshell, the successful application of the middle graph distillation method on GNNs in various graph analysis tasks shows the importance of middle knowledge to GKD technology.

\begin{table*}[htbp]
    \caption{A taxonomy of \zty{Self-Knowledge Distillation based \zty{Graph-based Knowledge Distillation} (SKD)}.}
    \centering
    \resizebox{1.0\linewidth}{!}{
    \begin{tabular}{c|ccc|c|c|c}
    \toprule
        \multirow{2}{*}{Method} & \multicolumn{3}{c|}{Distillation Location}                                                      & \multirow{2}{*}{Distance Measurement} & \multirow{2}{*}{Task}    & \multirow{2}{*}{Application}                        \\ \cline{2-4}
                                & \multicolumn{1}{c|}{Output Layer} & \multicolumn{1}{c|}{Intermediate Layer} & Constructed Graph &                                       &                          &                                                     \\ \hline
        LinkDist~\cite{LinkDist}                & \multicolumn{1}{c|}{\checkmark}            & \multicolumn{1}{c|}{}                   &                   & MSE                                   & Model compression        & Node classification                                 \\ \hline
        IGSD~\cite{IGSD}                    & \multicolumn{1}{c|}{\checkmark}            & \multicolumn{1}{c|}{}                   &                   & InfoCE                                & Graph-level task         & Graph classification, molecular property prediction \\ \hline
        GNN-SD~\cite{GNN-SD}                  & \multicolumn{1}{c|}{}            & \multicolumn{1}{c|}{\checkmark}                   &                   & KL, $L_2$                                & Relieve over-smoothing   & Node \& graph classification                        \\ \hline
        SDSS~\cite{SDSS}                    & \multicolumn{1}{c|}{\checkmark}            & \multicolumn{1}{c|}{\checkmark}                  & \checkmark                 & KL, MSE                               & Semi-supervised learning & Multitask node classification                       \\ \hline
        SAIL~\cite{SAIL}                    & \multicolumn{1}{c|}{\checkmark}            & \multicolumn{1}{c|}{}                   &                   & KL                                    & Unsupervised learning    & Node classification  \& clustering, link prediction \\ 
        \bottomrule
    \end{tabular}
    }
    \label{tab:skd-taxonom}
\end{table*}

\subsubsection{Constructed Graph Knowledge}
To further enrich and provide more general knowledge, GNNs will further learn the topological structure and node relationship information of the teacher model with the help of constructed graphs~\cite{GRL,GFL,HGKT,CPF,LSP,scGCN,MetaHG,G-CRD,HIRE}, so as to deeply explore the knowledge contained in the teacher model.

Among such methods, LSP [16] is the first knowledge distillation framework specially designed for homogeneous GNNs.     In particular, Yang \etal~\cite{LSP} develop a local structure preservation module LSP in GNN to capture graph topology information and realize the knowledge transfer from teacher to student model by minimizing the distance between distributions.     Inspired by LSP,  CPF~\cite{CPF}  is proposed, which designs the student model as a trainable combination of parametric label propagation and feature transformation modules so that students can benefit from prior knowledge based on structure and features in the teacher model.     Joshi \etal~\cite{G-CRD}   design a novel distillation framework G-CRD  for graph contrast learning representation based on LSP, which implicitly preserves the global topology based on the idea of contrastive learning to align the node embedding representation of teachers and students.     Song \etal~\cite{scGCN}  utilize a single-cell graph convolutional network model named scGCN, combine with knowledge distillation technology, to achieve effective knowledge transfer across different datasets. Qian \etal~\cite{MetaHG} leverage a meta-learning distillation framework MetaHG, which aims to jointly model structured relationships and unstructured content information on social media as heterogeneous graphs, and introduce meta-learning to transfer graph structure knowledge from the training task and effectively generalize it to the downstream test illegal drug trafficking task to solve the problem of label sparseness.         Ma \etal~\cite{GRL} propose a new multi-task knowledge distillation method (GRL) for graph-level representation learning, which presents graph metrics based on network theory as an auxiliary task to learn better graph representations through multi-task learning.         Yao \etal~\cite{GFL} present a graph few-shot learning model GFL, which learns a transferable metric space with the help of constructed auxiliary graphs o better capture global information.         Wang \etal~\cite{HGKT} perform a knowledge transfer method based on heterogeneous graphs (HGKT), which captures the inter-class and intra-class relationships with the help of structured heterogeneous graphs constructed and calculates the node representation of invisible classes by transferring the knowledge of the adjacent classes of invisible classes.

Furthermore, in addition to the knowledge distillation model for homogeneous graphs, Liu \etal~\cite{HIRE} also develop HIRE, a high-order relational knowledge distillation framework, especially for heterogeneous GNNs. by integrating node-level knowledge distillation and relational-level knowledge distillation, the correlation knowledge between individual node soft labels and different node types is distilled to the student model.    Most importantly, HIRE is a practical and general training method, which is suitable for arbitrary heterogeneous GNNs and homogeneous GNNs.  HIRE not only improves the performance and generalization ability of heterogeneous student models but also guarantees the node-level and relation-level knowledge extraction of heterogeneous GNNs.

In summary, the GKD methods have achieved impressive performance improvement on GNNs, and have become a new paradigm for GNNs distillation learning.  Since GKD has only attracted the attention of scholars lately, there are still many problems to be explored and urgently solved in this field, which can be found in Section~\ref{sec:Discussion}.

\subsection{\zty{Self-Knowledge Distillation based \zty{Graph-based Knowledge Distillation}}}
With the development of \zty{Graph-based Knowledge Distillation}, another type of SKD method has been proposed, which has greatly attracted the wide attention of scholars and has become one of the hot spots in current research. Therefore, this section will focus on summarizing the Self-Knowledge Distillation based Graph-based Knowledge Distillation (SKD) methods in GNNs and classify the SKD methods based on knowledge into three types: output layer, middle layer, and constructed graph according to the distillation location, as subdivided in TABLE~\ref{tab:skd-taxonom}.
(Note that only the most prominent forms of knowledge distillation are highlighted in the description of various \zty{Graph-based Knowledge Distillation} methods.)

\textbf{Similarities:}
For the SKD methods based on the output layer, middle layer, and constructed graph, each type of \zty{Graph-based Knowledge Distillation} algorithm is based on the extraction of knowledge at the same location. For the DKD methods based on the output layer, middle layer, and constructed graph, each type of \zty{Graph-based Knowledge Distillation} algorithm is based on knowledge extraction at the same location. They all follow the DKD model framework of Fig.~\ref{fig:frame-skd} and the graph distillation loss  $\zty{L_{Self}}$ paradigm of Eq.~\ref{eq:skd-loss}.

\textbf{Differences:}

The differences between each SKD method are detailed in TABLE~\ref{tab:skd-taxonom}. For instance, SAIL~\cite{SAIL}, GNN-SD~\cite{GNN-SD}, and SDSS~\cite{SDSS} use KL divergence for output layer, middle layer, and constructed graph knowledge, respectively. LinkDist~\cite{LinkDist} applies GKD using MSE distance for node classification and model compression. IGSD~\cite{IGSD} uses MSE to alleviate over-smoothing. GNN-SD~\cite{GNN-SD} employs KL and $L_2$ distance metrics for graph classification and molecular property prediction, respectively. SDSS~\cite{SDSS} utilizes KL and MSE distance metrics for multi-task node classification in downstream semi-supervised learning.

\subsubsection{Output Layer Knowledge}
Output layer knowledge-based is also the most basic and commonly used method of SKD, which is essential to extract the knowledge of label category-related knowledge contained in the pre-trained teacher model. In the graph learning task, especially in the semi-supervised or even self-supervised learning of the graph, it is difficult to obtain label data. Based on this, the researchers begin to try to apply the SKD method to GNNs. Among them, the most representative work is IGSD proposed by Zhang \etal~\cite{IGSD} in 2020, which iteratively performs Teacher-Student distillation by discriminating examples of augmented views of graph instances.

In contrast with traditional knowledge distillation, IGSD is based on the framework of contrastive learning, combined with \zty{Self-Knowledge Distillation} distillation technology so that the teacher network and the student network can be trained at the same time, thereby enhancing the graph representation ability.   Later, Luo \etal~\cite{LinkDist} propose the LinkDist method, aiming to make MLP achieve or even exceed the expressiveness of GCN models in graph classification tasks by distilling knowledge from the edges of graph networks.   IGSD and LinkDist prove the superiority of the output layer knowledge method and show the great potential of the output layer knowledge distillation method based on SKD on GNN.

\subsubsection{Middle Layer Knowledge}
Although the knowledge based on the output layer is simple and effective, the expression ability of GNNs is limited if it only relies on the output supervised information. Therefore, researchers try to explore the rich information contained in the convolutional layer of the GNNs, hoping to mine the knowledge with more ability to express node features. The representative methods include GNN-SD~\cite{GNN-SD} and SAIL~\cite{SAIL}.

Considering that the two-stage T-S distillation method is time-consuming and the performance of the student model is limited by the teacher model, Chen \etal~\cite{GNN-SD} utilize the GNN-SD method to replace the traditional two-stage GNN distillation method to distill middle layer knowledge from shallow to deep for alleviating the problem of over-smoothing faced by GNN. Specifically, GNN-SD develops the Neighborhood Discrepancy Rate (NDR)  to quantify the non-smoothness of the shallow embedding of GNN and refine this knowledge into the deep representation of GNN. Meanwhile, on the basis of NDR, GNN-SD also designs an adaptive discrepancy retaining (ADR) regularizer to enhance the transferability of knowledge and maintain high neighborhood deviation between GNN layers. It is found that GNN-SD can effectively alleviate the problem of over-smoothing and can also significantly reduce the training cost of two-stage T-S knowledge distillation.

Another representative work is the SAIL framework presented by Yu \etal~\cite{SAIL} in 2022. SAIL contains two complementary \zty{Self-Knowledge Distillation} distillation modules, namely Intra-distill and Inter-distillation modules, which iteratively use the smoothing node feature of the GNN intermediate layer to correct the GNN shallow representation through Intra-distill and Inter-distill. Experimental results show that SAIL is helpful for learning shallow GNNs with strong competitiveness, which is better than the GNNs obtained by current supervised or unsupervised training.

Overall, the preliminary research results of GNN-SD and SAIL reveal a promising method for realizing GKD.

\subsubsection{Constructed Graph Knowledge}
Besides the knowledge forms of the output layer and the middle layer, researchers also pay attention to the knowledge based on the constructed graph. Compared with the previous two kinds of knowledge, the constructed graph knowledge can extract the correlation information between the node features in the model, and the model knowledge can be fully mined and utilized to open up a new perspective for the research of SKD.

Ren \etal~\cite{SDSS} argue that the mismatch between the graph structure and the label affects the performance of the model. A multi-task self-distillation frame SDSS is proposed, which injects self-supervised learning and \zty{Self-Knowledge Distillation} distillation into a graph convolution network. Also, it solves the mismatch between structural and label by digging into the information in graphs and labels. In particular, the SDSS method uses four pre-task self-supervised learning to extract different levels of graph similarity information to facilitate local feature aggregation of the graph convolution network.   Experimental results show that the method has achieved impressive performance improvement in several classical graph convolution models.

The success of SDSS shows that self-supervision and \zty{Self-Knowledge Distillation} distillation are well integrated with the GNN framework, which provides new ideas and directions for subsequent researchers.

\section{Experiment}\label{sec:Experiment}
In this section, we compare and analyze the \zty{Graph-based Knowledge Distillation} method for deep neural networks (DKD), the \zty{Graph-based Knowledge Distillation} method for graph neural networks (GKD), and the \zty{Self-Knowledge Distillation} based \zty{Graph-based Knowledge Distillation} (SKD) respectively for comparative analysis. Firstly, we describe commonly used datasets. Then, we introduce the experimental setting. Finally, the experimental results of the \zty{Graph-based Knowledge Distillation} method are analyzed.

\subsection{Datasets}
To compare the experimental effects before and after using DKD, two datasets commonly used in DNNs are selected, including CIFAR-10~\cite{CIFAR} and CIFAR-100~\cite{CIFAR}. The dataset details are illustrated in TABLE~\ref{tab:data-dnn} below.

\begin{table}[htbp]
  \caption{The common datasets and statistics of DKD. $\vert Total \vert$ is the total number of images in the dataset. $\vert Train \vert$ is the number of training sets in the dataset. $\vert Test\vert$ is the number of test sets in the dataset.}
  \centering
  \resizebox{0.8\linewidth}{!}{
  \begin{tabular}{lrrrr}
    \toprule
    Dataset            & $\vert Total \vert$ & $\vert Train \vert$ & $\vert Test \vert$ & Class \\
    \midrule
    Cora      & 60000           & 50000         & 10000  & 10 \\
    Citeseer  & 60000           & 50000          & 10000  & 100 \\
    \bottomrule
  \end{tabular}
  }
  \label{tab:data-dnn}
\end{table}
\begin{itemize}
  \item \textbf{CIFAR-10:} is a small color image dataset with a total of 60000 color images, divided into 10 classes with 6000 images per category. Among them, there are 50000 images for training and another 10000 images for testing.
  \item \textbf{CIFAR-100:} consists of 60000 color images in 100 classes, each containing 600 images. Each class has 500 training images and 100 test images.
\end{itemize}

To compare the experimental results of GKD methods, seven datasets commonly used in GNNs, including Cora, Citeseer, Pubmed, Amazon-Photo (A-P), Amazon-Computers (A-C), Coauthor-Physics (Physics), Coauthor-CS (CS). The specific information of the dataset is drawn in TABLE~\ref{tab:data-gnn}.

\begin{table}[htbp]
  \caption{The common datasets and statistics of GKD and SKD. $\vert V \vert$ is the number of nodes in the dataset. $\vert E \vert$ is the number of edges.}
  \centering
  \resizebox{0.8\linewidth}{!}{
  \begin{tabular}{lrrrr}
    \toprule
    Dataset            & $\vert V \vert$ & $\vert E \vert$ & Feature & Class \\
    \midrule
    Cora      & 2708            & 5278          & 1443  & 7 \\
    Citeseer  & 3327            & 4552          & 3703  & 6 \\
    Pubmed  & 19717             & 44324         & 500   & 3 \\
    A-P     & 7650             & 119043         & 745  & 8 \\
    A-C     & 13752            & 245778         & 767   & 10\\
    Physics     & 34493     & 247962          & 8415  & 5 \\
    CS    & 18333           & 81894           & 6805 & 15  \\  
    \bottomrule
  \end{tabular}
  }
  \label{tab:data-gnn}
\end{table}
\begin{itemize}
   \item \textbf{Cora:} is a benchmark citation dataset composed of machine learning papers~\cite{CoraCiteseerPubmed}.  Nodes represent papers, and edges represent citation relationships.  Each node has a 1433-dimensional feature, and the class label indicates the research field to which each paper belongs.  The task is to classify the paper into different fields according to the citation network.
  \item \textbf{Citeseer:} is another commonly used benchmark citation dataset~\cite{CoraCiteseerPubmed}.  Each node represents a paper, each edge represents the citation relationship between two papers, the node feature dimension is 3703, there are six class labels, and the task is to predict the category of a publication.
  \item \textbf{Pubmed:} is also a citation network~\cite{CoraCiteseerPubmed}, containing 19717 nodes and 44324 edges, where nodes represent diabetes-related papers and edges represent the relationships between referenced papers.  The node is characterized by TF/IDF weighted word frequency with 500 dimensions.  The category label has three categories, and the task is to predict the type of diabetes in the paper.
  \item \textbf{A-P and A-C:} are the product purchase network of Amazon~\cite{Amazondata}.  Nodes represent goods, and edges indicate that the two are often purchased together.  The node feature is represented by a bag of words for a product review, and the task is to predict the category of the item.
  \item \textbf{Physics and CS:}  are commonly used citation networks extracted from the Microsoft Academic Graph from the KDD Cup 2016 Challenge~\cite{Amazondata}.  The node indicates the author, and the edge indicates whether the author is in a cooperative relationship.  Node features are represented by the keywords of each author's published paper, and the category label indicates the research area of each author.  Given the keywords for each author's paper, the task is to divide the authors into their respective fields of study.
\end{itemize}

For simplicity, to compare various SKD methods, three representative datasets, including Cora, Citeseer, and Pubmed, are selected. The dataset details are demonstrated in TABLE~\ref{tab:data-gnn}.

\subsection{Experiment Setting}
\textbf{DKD}. For simplicity, the representative Resnet-20~\cite{Resnet} model is selected as the framework of the deep neural network model to test the performance of the \zty{Graph-based Knowledge Distillation} method on the image classification task of the above two datasets (CIFAR-10 and CIFAR-100). We utilize the Accuracy metric. In order to compare the distillation effect, we choose the classical KD and IRG, RKD, and CC, which are three commonly used DKD methods. The specific classification effect can be found in TABLE~\ref{tab:result-dkd}. (Note: Bold indicates optimal performance, underline indicates suboptimal performance, and italic indicates performance degradation).

\textbf{GKD}. In the comparison experiment of the node classification task, the most representative model of GNNs (i.e., GCN~\cite{GCN}, GAT~\cite{GAT}, and SAGE~\cite{SAGE}) are selected on the seven datasets shown in TABLE~\ref{tab:data-gnn}. We leverage the classical KD and CPF distillation methods. We select F1-Micro and F1-Macro metrics to evaluate the distillation effect. Specific classification results can be found in TABLE~\ref{tab:result-clf}. (Note: Bold indicates optimal performance, underline indicates suboptimal performance, and italic indicates performance degradation). Similarly, in terms of clustering tasks, we adopt NMI and ARI clustering metrics and employ three GNNs (GCN, GAT, and SAGE) to apply KD and CPF knowledge distillation methods. See the next section for specific experimental results found in TABLE~\ref{tab:result-clu}. In addition, to quantitatively analyze the distillation effect, the experimental results of node visualization are also presented. Specifically, t-sne~\cite{t-sne} is performed on the node representation before and after the knowledge distillation of GCN, GAT, and SAGE models. The visualization results are summarized in Fig.~\ref{fig:vis-gcn}, Fig.~\ref{fig:vis-gat}, and Fig.~\ref{fig:vis-sage}.

\textbf{SKD}. 
For simplicity, we select the classical GCN model as the graph neural network model framework and take node classification as the task to test the distillation effect on Cora, Citeseer, and Pubmed datasets. Accuracy is selected as the classification metric. We apply the classic KD and LinkDist, SAIL, and SDSS \zty{Self-Knowledge Distillation} distillation methods. The specific classification results are depicted in TABLE~\ref{tab:result-skd}. (Note: Bold indicates optimal performance, underline indicates suboptimal performance, and italic indicates performance degradation).

All the experiments done in this paper run on V100GPU of NVIDIA Tesla and are based on the DGL~\cite{dgl} graph library with Pytorch-1.6.0 version 0.6~\cite{pytorch}.

\subsection{Evaluation and Results}

\subsubsection{\zty{\zty{Graph-based Knowledge Distillation} for Deep Neural Networks}}
\textbf{Experimental \zty{results} of DKD:}
From TABLE~\ref{tab:result-dkd}, KD, IRG, RKD, and CC are all consistent and significantly improve the image classification effect of the Resnet-20 teacher model. Among them, KD has the best distillation performance in CIFAR-10, followed by CC. On the CIFAR-100 dataset, IRG performs best, improving the image classification performance of the teacher model from 0.6982 to 0.7037. KD achieves sub-optimal, which improves the image classification performance of the teacher model from 0.6982 to 0.7036. Although KD, IRG, RKD, and CC have different gain effects on the Resnet-20 model, their performance is not very different. To a certain extent, this reflects that the combination of DNNs and KD has reached a certain bottleneck in the improvement of the Resnet-20 model. New graph distillation learning paradigms can be explored to further enhance the distillation effect for DNNs, such as combining Adversarial Learning, Neural Architecture Search, Graph Neural Networks, and other new technologies. A detailed discussion of this part can be seen in Section~\ref{sec:newparadigm}.

\begin{table}[htbp]
    \caption{The effect comparison of DKD. Bold fonts indicate the best performance. Underlined fonts indicate suboptimal performance. Italics indicate performance degradation.}
    \centering
    \resizebox{.85\linewidth}{!}{
    \begin{tabular}{c|c|cc}
        \toprule
            Model   & Metric   & CIFAR-10        & CIFAR-100       \\ 
            \hline
            Teacher & Accuracy & 0.9237          & 0.6892          \\ 
            +KD     & Accuracy & \textbf{0.9330} & {\underline {0.7036}}    \\ 
            +IRG    & Accuracy & 0.9277          & \textbf{0.7037} \\ 
            +RKD    & Accuracy & 0.9272          & 0.6948          \\ 
            +CC     & Accuracy & {\underline {0.9301}}    & 0.6927          \\ 
        \bottomrule
    \end{tabular}
    }
    \label{tab:result-dkd}
\end{table}

\subsubsection{\zty{\zty{Graph-based Knowledge Distillation} for Graph Neural Networks}}
\textbf{Experimental \zty{results} of GKD:}

\textbf{Node classification.} As can be seen from the node classification results in TABLE~\ref{tab:result-clf}, the classification performance of GCN, GAT, and SAGE student variants on seven datasets such as Cora has been uniformly and significantly improved under the guidance of classical KD or CPF \zty{Graph-based Knowledge Distillation} algorithm. In particular, CPF is much better than that of classical KD under the framework of GCN and GAT models. Nevertheless, the student variant model of SAGE shows the opposite performance. Although KD is better than CPF in most cases under SAGE model, they are similar (KD$\approx$CPF), performing similarly. In addition, although CPF is an integrated method for knowledge distillation at multiple locations, its performance on some datasets is inferior to KD, such as the distillation performance of GCN model on Pubmed dataset, GAT model on CS dataset and SAGE model on Cora, Physics, and CS dataset are lower than KD. In brief, notwithstanding that both KD and CPF \zty{Graph-based Knowledge Distillation} methods significantly improved the node classification performance of GNN model, they could not maintain the best performance on Cora, Citeseer, Pubmed, A-P, A-C, Physics, and CS. Similarly, the distillation effect of the same algorithm on different GNN models is different. For instance, CPF can improve GCN performance from 80.6\% to 85.62\%, GAT model from 82.08\% to 85.76\%, and SAGE model from 79.80\% to 81.83\% (taking the F1-Micro metric of node classification as an example). This reflects that, when designing the \zty{Graph-based Knowledge Distillation} algorithm, it is necessary to consider not only the combination of knowledge of different distillation positions but also the applicability of GNN model and the universality of dataset. In other words, it is vital to design a powerful \zty{Graph-based Knowledge Distillation} algorithm that can be applied to arbitrary GNN models and can obtain SOTA (State-Of-The-Art) distillation effect on any dataset at the same time.

\begin{table*}[htbp]
  \caption{The node classification effect comparison of GKD. Bold fonts indicate the best performance. Underlined fonts indicate suboptimal performance. Italics indicate performance degradation.}
  \centering
  \resizebox{.8\linewidth}{!}{
  \begin{tabular}{llccccccc}
    \toprule
    Model                    & Metric       & Cora            & Citeseer        & Pubmed          & A-P             & A-C             & Physics         & CS \\
    \midrule
    \multicolumn{9}{c}{Node classification of graph distillation variants based on GCN}   \\
    \midrule
        \multirow{2}{*}{GCN}  & F1-Micro & 0.8096          & 0.7086          & 0.7912          & 0.6441          & 0.4727          & 0.9226          & 0.8918          \\ 
                      & F1-Macro & 0.7985          & 0.6789          & 0.7862          & 0.6374          & 0.4347          & 0.9002          & 0.8698          \\ \hline
        \multirow{2}{*}{+KD}  & F1-Micro & {\underline {0.8276}}    & {\underline {0.7364}}    & \textbf{0.8032} & {\underline {0.7982}}    & {\underline {0.6759}}    & {\underline {0.9322}}    & {\underline {0.9105}}    \\ 
                      & F1-Macro & {\underline {0.8163}}    & \textbf{0.6916} & \textbf{0.7964} & {\underline {0.7859}}    & {\underline {0.6672}}    & {\underline {0.9107}}    & \textbf{0.8862} \\ \hline
        \multirow{2}{*}{+CPF} & F1-Micro & \textbf{0.8562} & \textbf{0.7530} & {\underline {0.6811}}    & \textbf{0.9313} & \textbf{0.8459} & \textbf{0.9476} & \textbf{0.9126} \\ 
                      & F1-Macro & \textbf{0.8261} & {\underline {0.6623}}    & {\underline {0.6716}}    & \textbf{0.9162} & \textbf{0.8456} & \textbf{0.9304} & {\underline {0.8860}}    \\ 
    \midrule
    
    \multicolumn{9}{c}{Node classification of graph distillation variants based on GAT} \\
    \midrule
        \multirow{2}{*}{GAT}  & F1-Micro          & 0.8208          & 0.7032          & 0.7722          & 0.8054          & 0.6559          & 0.9252          & 0.9066          \\ 
                      & F1-Macro          & 0.8116          & 0.6732          & 0.7684          & 0.7972          & 0.6244          & 0.9023          & 0.8824          \\ \hline
                      
        \multirow{2}{*}{+KD}  & F1-Micro          & {\underline {0.8410}}    & {\underline {0.7264}}    & {\underline {0.7848}}    & {\underline {0.8421}}    & {\underline {0.6688}}    & {\underline {0.9350} }   & \textbf{0.9090} \\ 
                      & F1-Macro & \textbf{0.8331} & \textbf{0.6818} & {\underline {0.7780}}    & {\underline {0.8273}}    & {\underline {0.6535}}    & {\underline {0.9136}}    & \textbf{0.8859} \\ \hline
                      
        \multirow{2}{*}{+CPF} & F1-Micro          & \textbf{0.8576} & \textbf{0.7541} & \textbf{0.7949} & \textbf{0.9158} & \textbf{0.8456} & \textbf{0.9407} & {\underline {0.9031}}    \\ 
                      & F1-Macro          & {\underline {0.8295}}    & {\underline {0.6692}}    & \textbf{0.7938} & \textbf{0.8981} & \textbf{0.8516} & \textbf{0.9219} & {\underline {0.8743}}    \\
    \bottomrule
   
    \multicolumn{9}{c}{Node classification of graph distillation variants based on SAGE} \\
    \midrule
    \multirow{2}{*}{SAGE} & F1-Micro          & 0.7980          & 0.7052          & 0.7844          & 0.8754          & 0.7660          & 0.9239          & 0.9203          \\ 
                      & F1-Macro          & 0.7862          & 0.6769          & 0.7824          & 0.8663          & 0.7599          & 0.9021          & 0.8999          \\ \hline
        \multirow{2}{*}{+KD}  & F1-Micro          & \textbf{0.8198} & {\underline {0.7202}}    & {\underline {0.7962}}    & {\underline {0.8853}}    & \textbf{0.7838} & \textbf{0.9379} & \textbf{0.9262} \\ 
                      & F1-Macro & \textbf{0.8100} & \textbf{0.6877} & {\underline {0.7935}}    & {\underline {0.8763}}    & {\underline {0.7821}}    & \textbf{0.9163} & \textbf{0.9062} \\ \hline
        \multirow{2}{*}{+CPF} & F1-Micro          & {\underline {0.8183}}    & \textbf{0.7365} & \textbf{0.8053} & \textbf{0.9249} & {\underline {0.7824}}    & {\underline {0.9363}}    & {\underline {0.9016}}    \\ 
                      & F1-Macro          & {\underline {0.7861}}    & {\underline {0.6148}}    & \textbf{0.8051} & \textbf{0.9067} & \textbf{0.8078} & {\underline {0.9135}}    & {\underline {0.8682}}    \\
    \bottomrule
  \end{tabular}
  }
  \label{tab:result-clf}
\end{table*}

\textbf{Node clustering.} Besides performing node classification tasks, we also conduct node clustering experiments on the seven datasets. Specific clustering results are mirrored in TABLE~\ref{tab:result-clu}. We found the gain effect of KD and CPF on GNN teacher model is quite different. Overall, under the guidance of KD, the clustering performance of GCN, GAT, and SAGE student models has been improved to some extent, but their distillation effect is not the same. Among them, it is known from TABLE~\ref{tab:result-clu} that the KD performs better than CPF on Cora, Pubmed, A-P, and CS datasets, while CPF is better than that of KD on Citeseer, A-C, and Physics. It can be observed that under the backbone of GAT and SAGE, KD is better than CPF on the whole and can even make the corresponding student model maintain the optimal performance under each dataset. In addition, an interesting experimental phenomenon is found in the knowledge distillation experiment. CPF will damage the performance of GNN model, which decreases the node clustering performance of the corresponding teacher model on all datasets, especially on GAT and SAGE models. For example, under the guidance of CPF, the teacher performance of CS dataset based on SAGE model was reduced from 79.88\% to 57.79\% (taking the NMI metric of node clustering as an example). It is very different from the CPF \zty{Graph-based Knowledge Distillation} algorithm in node classification. To some extent, it is still challenging to combine GNNs and graph distillation algorithms. How to better apply \zty{Graph-based Knowledge Distillation} to GNN still needs further exploration. We have an in-depth discussion and prospect regarding this part in Section~\ref{sec:Discussion}.

\begin{table*}[htbp]
  \caption{The node clustering effect comparison of GKD. Bold fonts indicate the best performance. Underlined fonts indicate suboptimal performance. Italics indicate performance degradation.}
  \centering
  \resizebox{.8\linewidth}{!}{
  \begin{tabular}{llccccccc}
    \toprule
    Model                    & Metric       & Cora            & Citeseer        & Pubmed          & A-P             & A-C             & Physics         & CS \\
    \midrule
    \multicolumn{9}{c}{Node clustering of graph distillation variants based on GCN}   \\
    \midrule
        \multirow{2}{*}{GCN}  & NMI          & 0.5568          & 0.4291                   & 0.3711          & 0.4235          & 0.3399          & 0.7029                   & 0.6736          \\ 
                      & ARI          & 0.5120          & {\underline {0.4241}}             & 0.4094          & 0.2619          & 0.2083          & 0.6806                   & 0.5336          \\ \hline
        \multirow{2}{*}{+KD}  & NMI          & \textbf{0.6011} & {\underline {0.4655}}             & \textbf{0.3874} & \textbf{0.5932} & {\underline {0.4578}}    & {\underline {0.7111}}             & \textbf{0.7145} \\ 
                      & \textbf{ARI} & \textbf{0.5933} & 0.4620                   & \textbf{0.4401} & \textbf{0.4655} & {\underline {0.2883}}    & {\underline {0.6890}}             & \textbf{0.6025} \\ \hline
        \multirow{2}{*}{+CPF} & NMI          & {\underline {0.5988}}    & \textbf{0.4714}          & \textit{0.1935} & {\underline {0.5888}}    & \textbf{0.5299} & \textbf{0.7519}          & \textit{0.5299} \\
                      & ARI          & {\underline {0.5801}}    & \textit{\textbf{0.4721}} & \textit{0.1214} & {\underline {0.3872}}    & \textbf{0.2985} & \textit{\textbf{0.8228}} & \textit{0.2985} \\ 
    \midrule
    
    \multicolumn{9}{c}{Node clustering of graph distillation variants based on GAT} \\
    \midrule
        \multirow{2}{*}{GAT}  & NMI          & 0.6056          & 0.4297          & 0.3626                   & 0.6545          & 0.4975          & 0.7669          & 0.7531          \\ 
                      & ARI          & 0.5634          & 0.4257          & 0.3910                   & 0.5311          & 0.4018          & 0.8391          & 0.6889          \\ \hline
        \multirow{2}{*}{+KD}  & NMI          & \textbf{0.6145} & {\underline {0.4550}}    & {\underline {0.3754}}             & \textbf{0.6814} & \textbf{0.5567} & \textbf{0.7711} & \textbf{0.7719} \\ 
                      & ARI & \textbf{0.5799} & \textbf{0.4449} & {\underline {0.4169}}             & \textbf{0.5975} & \textbf{0.4767} & \textbf{0.8506} & \textbf{0.7930} \\ \hline
        \multirow{2}{*}{+CPF} & NMI          & {\underline {0.6066}}    & \textbf{0.4551} & \textbf{0.4021}          & \textit{0.5113} & {\underline {0.4981}}    & \textit{0.6147} & \textit{0.5850} \\ 
                      & \textit{ARI} & \textit{0.5109} & \textit{0.4177} & \textit{\textbf{0.4266}} & \textit{0.2884} & \textit{0.2994} & \textit{0.5654} & \textit{0.4371} \\
    \bottomrule
   
    \multicolumn{9}{c}{Node clustering of graph distillation variants based on SAGE} \\
    \midrule
        \multirow{2}{*}{SAGE} & NMI          & 0.5707          & 0.4374                   & 0.4083          & 0.6870          & 0.5380          & 0.7641          & 0.7988          \\ 
                      & ARI          & 0.5433          & 0.4457                   & 0.4564          & 0.5813          & 0.3686          & 0.8238          & 0.7509          \\ \hline
        \multirow{2}{*}{+KD}  & NMI          & \textbf{0.5921} & {\underline {0.4618}}             & \textbf{0.4177} & \textbf{0.7010} & \textbf{0.5775} & \textbf{0.7854} & \textbf{0.8149} \\ 
                      & ARI & \textbf{0.5825} & {\underline {0.4597}}             & \textbf{0.4632} & \textbf{0.6175} & \textbf{0.4499} & \textbf{0.8643} & \textbf{0.8397} \\ \hline
        \multirow{2}{*}{+CPF} & NMI          & \textit{0.4892} & \textbf{0.4737}          & \textit{0.3598} & \textit{0.4666} & \textit{0.4808} & \textit{0.6323} & \textit{0.5779} \\ 
                      & \textit{ARI} & \textit{0.2965} & \textit{\textbf{0.4805}} & \textit{0.3724} & \textit{0.2803} & \textit{0.3104} & \textit{0.5655} & \textit{0.3894} \\ 
    \bottomrule
  \end{tabular}
  }
  \label{tab:result-clu}
\end{table*}

\begin{figure*}[htbp]
    \centering
  \includegraphics[width=1.0\textwidth]{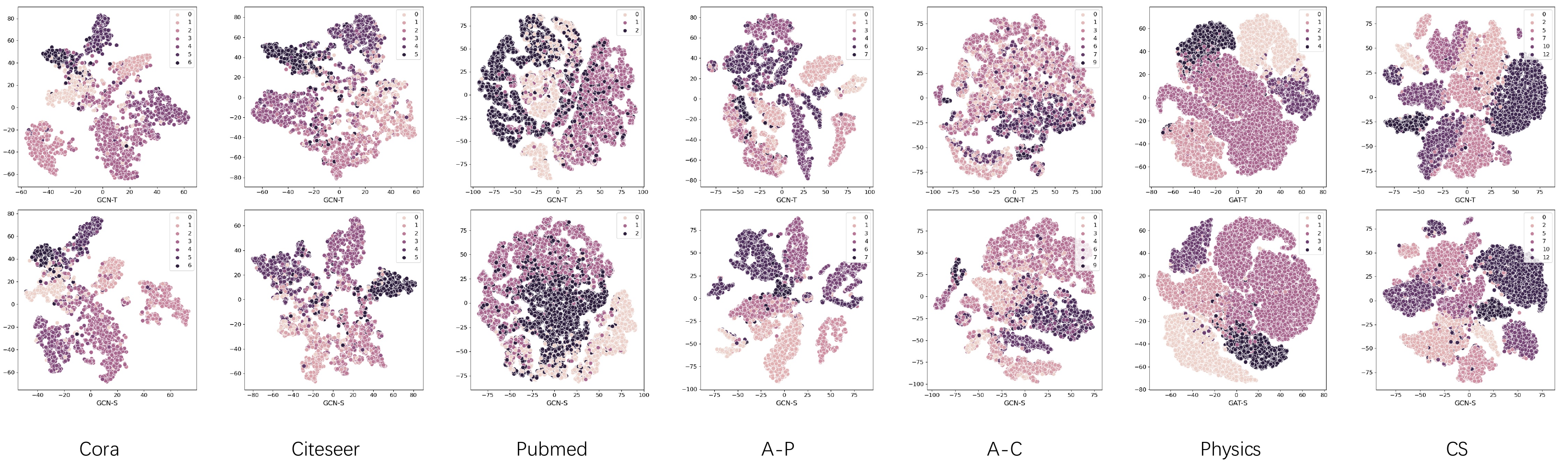}
  \caption{The node visualization of the GCN teacher model and its student variant.}
  \label{fig:vis-gcn}
\end{figure*}

\begin{figure*}[htbp]
    \centering
  \includegraphics[width=1.0\textwidth]{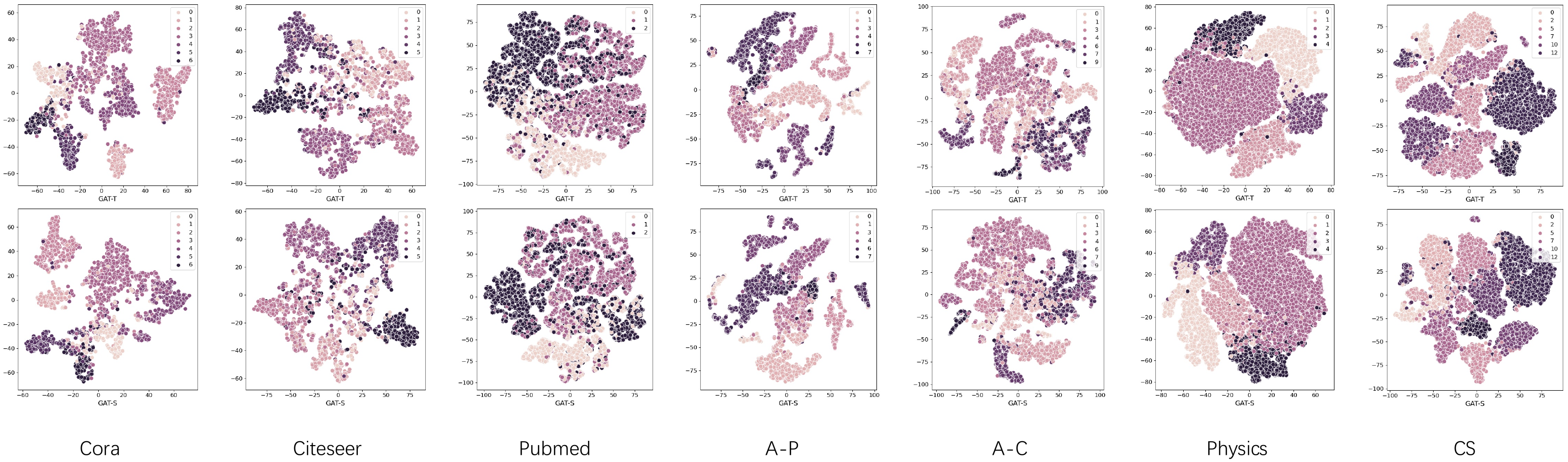}
  \caption{The node visualization of the GAT teacher model and its student variant.}
  \label{fig:vis-gat}
\end{figure*}

\begin{figure*}[htbp]
    \centering
  \includegraphics[width=1.0\textwidth]{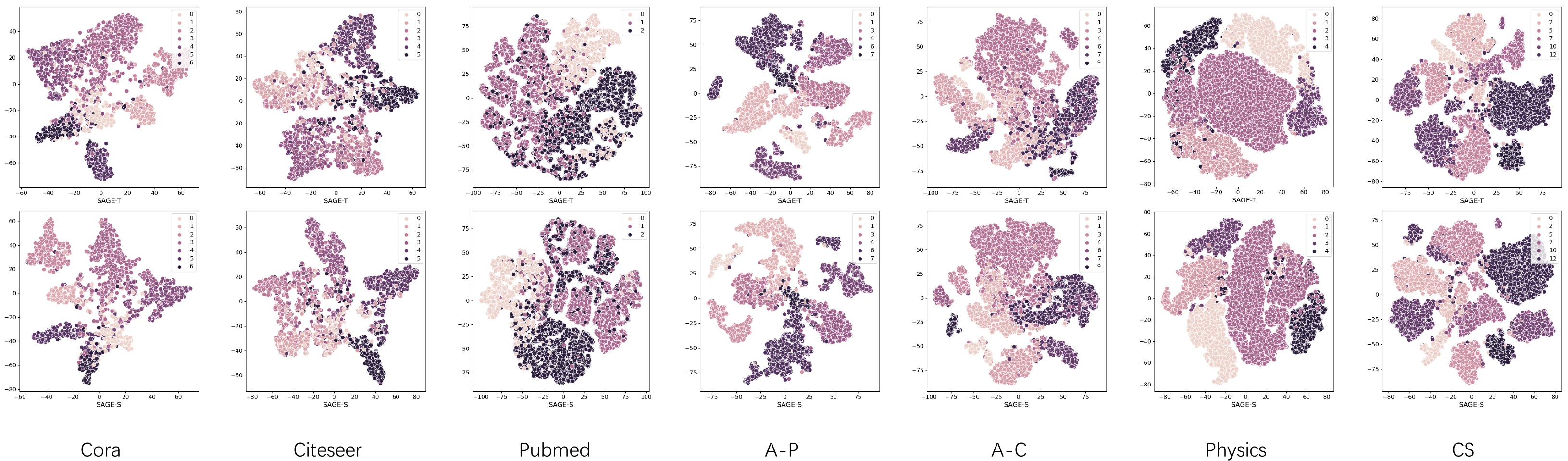}
  \caption{The node visualization of the SAGE teacher model and its student variant.}
  \label{fig:vis-sage}
\end{figure*}

\textbf{Node visualization.} In addition to node classification and clustering quantitative analysis of the knowledge distillation effect, we furthermore conduct a qualitative analysis of node visualization. The visualization results of GCN, GAT, and SAGE variants are illustrated in Fig.~\ref{fig:vis-gcn}, Fig.~\ref{fig:vis-gat}, and Fig.~\ref{fig:vis-sage}. The first line is the visualization effect of the corresponding teacher model, and the second line is the visualization result of the corresponding student model under the guidance of \zty{Graph-based Knowledge Distillation}. Through these three figures, it can be clearly found that the student clustering effect of GCN, GAT, and SAGE has been improved, where the boundary interval between different types of nodes has become larger, and the nodes of the same kind have been more closely clustered. Notwithstanding that the \zty{Graph-based Knowledge Distillation} algorithm can improve the node representation ability of the GNN model and make the classification interface of different category labels clearer, their effects are different on different datasets. This indicates that the current research of GKD on GNN still has great potential, and there are still many problems worthy of further research and exploration, which we have discussed and prospected in depth in the following section~\ref{sec:Discussion}.

\subsubsection{\zty{Self-Knowledge Distillation based \zty{Graph-based Knowledge Distillation}}}
\textbf{Experimental \zty{results} of SKD:}
It can be seen from the classification results of TABLE~\ref{tab:result-skd} that the classification performance of GCN student variants on all datasets has been improved more or less under the guidance of KD or LinkDist, SAIL, and SDSS distillation methods. Particularly, the distillation effect of SAIL and SDSS is much better than that of KD, greatly improving the performance of GCN teacher model. Whereas, LinkDist is not the same, whose distillation performance is lower than that of KD on the Cora and Pubmed datasets. Meanwhile, it is found that KD and LinkDist will reduce the node classification performance of their corresponding teacher models on Cora and Pubmed datasets. In brief, it is no doubt that KD and SKD methods can improve the node classification performance of GNN models, but they cannot maintain the best performance on all datasets in the meantime. This also reflects that it is necessary to consider whether the selection of distillation mode is appropriate or not when designing the \zty{Graph-based Knowledge Distillation} algorithm. Moreover, the determination of the distillation position and selection for distillation distance measurement also affects the distillation effect. A detailed discussion can be seen in Section~\ref{sec:Discussion}. In addition, SKD methods are currently in the preliminary exploration stage, and the research methods are few and lack theoretical support. Therefore, it is necessary to fully explore the theoretical analysis of interpretability behind the \zty{Graph-based Knowledge Distillation} method so as to design an efficient \zty{Graph-based Knowledge Distillation} method. For a detailed discussion, see Section~\ref{sec:interpretability}.

\begin{table}[t]
    \caption{The effect comparison of SKD. Bold fonts indicate the best performance. Underlined fonts indicate suboptimal performance. Italics indicate performance degradation.}
    \centering
    \resizebox{1.0\linewidth}{!}{
    \begin{tabular}{c|c|ccc}
        \toprule
            Model     & Metric   & Cora            & Citeseer        & Pubmed          \\ 
            \hline
            Teacher   & Accuracy & 0.8183          & 0.6762          & 0.7859          \\ 
            +KD       & Accuracy & \textit{0.8005} & 0.6821          & \textit{0.7571} \\ 
            +LinkDist & Accuracy & \textit{0.7572} & 0.7119          & \textit{0.7484} \\ 
            +SAIL     & Accuracy & {\underline{0.8463}}    & {\underline{0.7424}}    & \textbf{0.8381} \\ 
            +SDSS     & Accuracy & \textbf{0.8600} & \textbf{0.7613} & {\underline{0.8221}}    \\ 
        \bottomrule
    \end{tabular}
    }
    \label{tab:result-skd}
\end{table}

In short, the three types of \zty{Graph-based Knowledge Distillation} methods can improve the performance of CNN/GNN models by virtue of their advantages of model compression, model enhancement, simplicity, and efficiency, which are successfully applied in the recommendation system and other practical application scenarios. Although these methods have achieved good results, they still have certain shortcomings. Next section, we make a foreground on the improved direction of \zty{Graph-based Knowledge Distillation}: (1) Determination of distillation position; (2) Choice of distillation mode; (3) Selection for distillation distance measurement; (4) Theoretical analysis of interpretability; and (5) A new paradigm of \zty{Graph-based Knowledge Distillation}.

\section{Applications}\label{sec:Applications}
Knowledge distillation has attracted much attention from academia and industry it was proposed. With the development of knowledge distillation technology, \zty{Graph-based Knowledge Distillation} has achieved excellent performance in model compression and model enhancement, which has very important applications and broad prospects in Computer Vision, Natural Language Processing, Recommendation Systems, and other fields. In this section, this paper summarizes several common application scenarios of \zty{Graph-based Knowledge Distillation} in TABLE~\ref{tab:application} that are critical to better understanding and using \zty{Graph-based Knowledge Distillation} techniques. For instance, the most commonly adopted is T-S distillation, and the combination of KD and GNN is also a noteworthy research work in the future.

\begin{table*}[htbp]
    \caption{Taxonomy of the applications of \zty{Graph-based Knowledge Distillation} methods.}
    \centering
    \resizebox{1.0\linewidth}{!}{
    \begin{tabular}{c|c|c|cc}
        \toprule
        Field                                        & Problem                     & Backbone & Algorithm                                                                                                                              \\ \hline
        \multirow{8}{*}{Computer Vision}             & Image classification        & CNN       & \begin{tabular}[c]{@{}c@{}}HKD~\cite{HKD}, GKD~\cite{GKD}, SPKD~\cite{SPKD}, HKDIFM~\cite{HKDIFM},\\ KDExplainer~\cite{KDExplainer}, TDD~\cite{TDD}\end{tabular} \\ \cline{2-4} 
                                                     & Image recognition           & CNN       & KTG~\cite{KTG}, MHGD~\cite{MHGD}, IRG~\cite{IRG}                                                                                            \\ \cline{2-4} 
                                                     & Robot localization          & GNN       & GCLN~\cite{GCLN}                                                                                                                       \\ \cline{2-4} 
                                                     & Object detection            & CNN       & DOD~\cite{DOD}, GD~\cite{GD}                                                                                                           \\ \cline{2-4} 
                                                     & Video classification        & CNN       & BAF~\cite{BAF}                                                                                                                        \\ \cline{2-4} 
                                                     & Event prediction            & GNN       & EGAD~\cite{EGAD}                                                                                                                     \\ \cline{2-4} 
                                                     & Person re-identification    & CNN       & GCMT~\cite{GCMT}, CC~\cite{CC}                                                                                                           \\ \cline{2-4} 
                                                     & Road marking                & CNN       & IntRA-KD~\cite{IntRA-KD}                                                                                                                   \\ \hline
        \multirow{5}{*}{Natural Language Processing} & Visual dialogue             & CNN       & CAG~\cite{CAG}                                                                                                                       \\ \cline{2-4} 
                                                     & Relation extraction         & CNN       & DKWISL~\cite{DKWISL}                                                                                                                     \\ \cline{2-4} 
                                                     & Video captioning            & CNN       & SPG~\cite{SPG}                                                                                                                        \\ \cline{2-4} 
                                                     & Machine translation         & CNN       & LAD~\cite{LAD}                                                                                                                       \\ \cline{2-4} 
                                                     & Metric learning             & CNN       & RKD~\cite{RKD}                                                                                                                        \\ \hline
        \multirow{4}{*}{Recommender System}          & Incremental learning        & GNN       & LWC-KD~\cite{LWC-KD}                                                                                                                    \\ \cline{2-4} 
                                                     & Collaborative filtering     & CNN       & DGCN~\cite{DGCN}                                                                                                                      \\ \cline{2-4} 
                                                     & Cold start                  & GNN       & PGD~\cite{PGD}                                                                                                                      \\ \cline{2-4} 
                                                     & Tail generalization         & GNN       & Cold Brew~\cite{ColdBrew}                                                                                                                 \\ \hline
        \multirow{4}{*}{Multi-task Learning}         & Transfer learning           & CNN       & IEP~\cite{IEP}                                                                                                                       \\ \cline{2-4} 
                                                     & Image recognition           & GNN       & GRL~\cite{GRL}                                                                                                                      \\ \cline{2-4} 
                                                     & Image recognition           & CNN       & MHGD~\cite{MHGD}                                                                                                                       \\ \cline{2-4} 
                                                     & Self knowledge distillation & GNN       & SDSS~\cite{SDSS}                                                                                                                      \\ \hline
        \multirow{2}{*}{Zero-Shot Learning}          & Data-free distillation      & GNN       & GFKD~\cite{GFKD}                                                                                                                      \\ \cline{2-4} 
                                                     & Model enhancement           & GNN       & HGKT~\cite{HGKT}                                                                                                                     \\ 
        \bottomrule
    \end{tabular}
    }
    \label{tab:application}
\end{table*}

\subsection{Computer Vision}
As an effective model compression/model enhancement technique, \zty{Graph-based Knowledge Distillation} is widely used in different fields of artificial intelligence, especially in Computer Vision. In recent years, a variety of \zty{Graph-based Knowledge Distillation} algorithms have been proposed for different visual tasks. Among them, \zty{Graph-based Knowledge Distillation} is mainly applied in the downstream task of image classification~\cite{HKD,GKD,SPKD,HKDIFM,KDExplainer,TDD} to achieve model enhancement, interpretability, model compression, and other goals. \zty{Graph-based Knowledge Distillation} also has a very important application in image recognition~\cite{KTG,MHGD,IRG}. By constructing auxiliary graphs as a carrier of knowledge, the relationship knowledge between samples in the teacher model is extracted into the student model, which further improves the performance of the student model. In addition, in unsupervised learning scenarios, graph distillation is utilized in pedestrian reidentification~\cite{CC,GCMT} modeling in computer vision. More importantly, as demonstrated in TABLE~\ref{tab:application}, \zty{Graph-based Knowledge Distillation} can also be acceptable to visual tasks such as object detection~\cite{DOD,GD}, robot localization~\cite{GCLN}, video classification~\cite{BAF}, event prediction~\cite{EGAD}, and road marking~\cite{IntRA-KD}.

\subsection{Natural Language Processing}
Natural Language Processing (NLP) is an important branch of computer science and artificial intelligence, which is one of the current hot research fields. NLP models are developing rapidly, from RNN, Transformer, ELMo, GPT, and BERT to now GPT-3, whose model structure and parameters have become more and more complex and large, which seriously hinders the deployment and training of language models. The emergence of knowledge distillation provides an effective lightweight and deep language model knowledge transfer method, which can solve the problem of language model deployment simply and efficiently, and has become a research hotspot in the field of NLP. Nowadays, more and more graph distillation work has been proposed to deal with NLP problems, including visual dialogue~\cite{CAG}, machine translation~\cite{LAD}, relationship extraction~\cite{DKWISL}, etc., as depicted in TABLE~\ref{tab:application}.

\subsection{Recommendation System}
Recommendation System, as the name suggests, models user preferences based on user attributes, historical behavior, and so on to produce recommendations that users prefer. With the rapid development of Deep Learning, the model structure of recommended system becomes more and more complex, network depth becomes deeper and deeper, and model parameters become more and more numerous. Equally, in the recommended system field, it is also faced with the problem that model calculation is expensive and cannot be run on mobile or embedded devices. To solve the contradiction between model effect and response speed, \zty{Graph-based Knowledge Distillation} came into being. Using \zty{Graph-based Knowledge Distillation} technology, the rich knowledge in the pre-trained powerful teacher model can be distilled into the lightweight student model of recommended online to enhance the generalization ability of the student model of the recommendation system and achieve the goal of easy deployment of the recommendation model. In addition, the combination of graph distillation and the recommended system can be used to solve cold start~\cite{PGD}, tail generalization~\cite{ColdBrew}, incremental learning~\cite{LWC-KD}, and other problems. TABLE~\ref{tab:application} lists the representative work combining \zty{Graph-based Knowledge Distillation} and recommended systems.

\subsection{Multi-task Learning}
Apart from being widely used in the above-mentioned Computer Vision, Natural Language Processing, and Recommended Systems, \zty{Graph-based Knowledge Distillation} is also leveraged in combination with other emerging technologies such as Graph Neural Networks, Transfer Learning, and Multi-task Learning. Specifically, Lee \etal~\cite{IEP} propose an interpretable embedding process (IEP) knowledge distillation method based on principal component analysis to explain and understand the process of embedding representation of deep neural network models. Ma \etal~\cite{GRL} employ constructed graphs to control teachers’ knowledge transfer and learn better graph representations by using graph metrics based on graph theory as auxiliary tasks through Multi-task Learning. Lee \etal~\cite{MHGD} leverage multi-head attention to extract the knowledge of the teacher's embedding process and make the student model have relational inductive bias ability through Multi-Task Learning. Ren \etal~\cite{SDSS} combine \zty{Self-Knowledge Distillation} distillation and multi-task Learning and proposed a two-phase training distillation framework. TABLE~\ref{tab:application} has sorted out the part of the current \zty{Graph-based Knowledge Distillation} work on this task for the reference of researchers.

\subsection{Zero-Shot Learning}
Likewise, \zty{Graph-based Knowledge Distillation} also performs excellently in Zero-Shot Learning. For instance, Deng \etal~\cite{GFKD} first develop a non-data distillation method tailored for GNN, using multivariate Bernoulli distributions to transfer the knowledge of graph structure from the pre-trained GNN, and introduce a gradient estimator to optimize the framework. Wang \etal~\cite{HGKT} present a knowledge transfer method HGKT based on heterogeneous graphs, which transfer knowledge from visible classes to new invisible classes by means of structured heterogeneous graphs constructed to represent the relationship between data, so as to solve the problem of classification of instances of visible and invisible classes. The mainstream methods used in this part of the work are given in TABLE~\ref{tab:application}.

\section{Discussions and Prospects}\label{sec:Discussion}
As a knowledge transfer technology, \zty{Graph-based Knowledge Distillation} improves the performance of deep neural networks and emerging graph neural networks with its advantages of model compression, model enhancement, simplicity, and efficiency, and has been successfully applied in practical business scenarios such as recommendation systems. Notwithstanding that \zty{Graph-based Knowledge Distillation} has achieved satisfactory performance and has become a popular research area, it still has many problems that need attention and is worthy of further exploration. In view of the shortcomings of current \zty{Graph-based Knowledge Distillation} methods, several potential research directions of \zty{Graph-based Knowledge Distillation} are proposed in this section:

\subsection{Determination of distillation position}
Through the inductive analysis of graph distillation work, most of the existing graph distillation methods utilize different types of knowledge source combinations, including output layer, middle layer, and constructed graph knowledge. Yet, it is not clear which position of knowledge plays an important role. Especially for the knowledge of middle layers and constructed graph, some choose a convolutional layer in the middle, and some choose all convolutional layers, but which layer is specifically selected for distillation, there is little research at present. How to design a faster, more general-purpose, precision-assured graph distillation method that can model all types of knowledge sources simultaneously remains challenging. Especially the relationship among these three knowledge sources and how they interact and influence each other is essential for the rational utilization of graph structure data information and the full mining of knowledge. Determination of distillation position is the focus of future research in the \zty{Graph-based Knowledge Distillation} domain.

\subsection{Choice of distillation mode}
The two popular graph distillation methods are T-S distillation mode and \zty{Self-Knowledge Distillation} distillation mode. Because of its flexibility, controllability, and ease, T-S distillation method is suitable for large-scale complex teacher model compression. The model \zty{Self-Knowledge Distillation} distillation method is widely leveraged in downstream business scenarios with large overhead because of its simple structure and efficient training efficiency. Whereas, there are still deficiencies in these two distillation methods: T-S is complex and time-consuming in training; \zty{Self-Knowledge Distillation} lacks theoretical support and is confined to problem scenarios with the comparable performance of teacher and student models. Meanwhile, there is currently a lack of comparative studies on the distillation methods of the two. Therefore, it is necessary to study how the choice of distillation mode affects the effectiveness of KD and how to design an efficient distillation framework.

\subsection{Selection for distillation distance measurement}
The performance of graph distillation is inseparable from the selection of distance metric function in training loss. Because knowledge distillation extracts knowledge distillation from the teacher model into the student model, the effect of this knowledge transfer is reflected in the design of the loss function in the model training. That is, it can only be demonstrated by evaluating the proximity of node/node characteristics between the student model and the teacher model. Therefore, it is crucial to design a good distillation loss function. However, there are various selection methods of the loss function, such as KL, MSE, InfoCE, etc., but there is still no conclusion as to which loss function should be selected in the graph distillation process to better guide the model training process of students. Therefore, how to select suitable distance measurement functions according to concrete scenes and problem is an urgent problem in graph distillation technology.

\subsection{Theoretical analysis of interpretability}\label{sec:interpretability}
Although a large number of \zty{Graph-based Knowledge Distillation} work has been successfully applied in various practical business scenarios, there are still few theoretical analyses on the interpretability of knowledge distillation. Recently, there have been some preliminary attempts at the interpretability of knowledge distillation, such as Yuan \etal~\cite{Labelsmooth} who explain the principle of KD from the perspective of label smoothing, arguing that the success of KD is not entirely due to the similarity information between teacher categories, but due to the regularization of soft targets. Nevertheless, this finding only applies to classification tasks, not to tasks without labels~\cite{li2020gan}. Cheng \etal~\cite{cheng2020explaining} interpret knowledge distillation in terms of quantitative knowledge, that is, by defining and quantifying the "amount of knowledge" of the features of the middle layer of the neural network, the success mechanism of the knowledge distillation algorithm is explained from the perspective of the expression ability of the neural network. Mobahi \etal~\cite{mobahi2020self} prove for the first time that \zty{Self-Knowledge Distillation} distillation plays the role of $L_2$ regularizer by fitting the training data in Hilbert space, thus providing some theoretical analysis for the \zty{Self-Knowledge Distillation} distillation method. Unfortunately, the interpretation of the middle layer, constructed graph, and other knowledge is very limited, and the distillation mechanism behind it remains unclear. Therefore, the mathematical principle behind the distillation effect has not been fully explored to a large extent, and the theoretical study of \zty{Graph-based Knowledge Distillation} method still deserves further exploration and attention, which has important guiding significance for exploring new efficient graph distillation methods.

\subsection{A new paradigm of \zty{Graph-based Knowledge Distillation} }\label{sec:newparadigm}
As \zty{Graph-based Knowledge Distillation} has shown impressive performance improvements in many tasks, lots of researchers have begun to try to combine it with existing new techniques, including Adversarial Learning, Neural Architecture Search, Graph Network Networks, Reinforcement Learning, Incremental Learning, Federated Learning, Quantization, and Prunning, etc. Knowledge distillation is used in combination with other technologies to derive a lot of practical applications. For example, knowledge distillation can be utilized as an effective strategy to defend against perturbations in deep neural networks~\cite{papernot2016distillation,ross2018improving} and can be adapted to solve data privacy and security problems~\cite{papernotsemi,wang2019private}. But these methods are still in the exploratory stage, and methods are not mature. Therefore, how to better combine knowledge distillation with other technologies is a valuable and meaningful future direction for the expansion of graph distillation to other uses and applications.

\section{Conclusion}\label{sec:Conclusion}
Based on the basic concept of graph data and knowledge distillation, the method of \zty{Graph-based Knowledge Distillation} is thoroughly combed in this paper. Firstly, on the basis of the design characteristics of the \zty{Graph-based Knowledge Distillation} algorithm, it can be divided into three categories: \zty{Graph-based Knowledge Distillation} for deep neural networks (DKD), \zty{Graph-based Knowledge Distillation} for graph neural networks (GKD), and \zty{Self-Knowledge Distillation} based \zty{Graph-based Knowledge Distillation} (SKD).  Secondly, it can be further subdivided into the output layer, middle layer, and constructed graph method based on the position of knowledge distillation. Then, the algorithm performance of the mainstream \zty{Graph-based Knowledge Distillation} method is compared experimentally. In addition, we summarize the critical application scenarios of \zty{Graph-based Knowledge Distillation} in various fields. Finally, we conclude and prospect the research direction of \zty{Graph-based Knowledge Distillation} learning recently. It is hoped that this paper can provide some insights for the researchers of graph representation learning and knowledge distillation to promote the sustainable development of this field.


\bibliographystyle{IEEEtran}

\bibliography{ref}

\begin{thebibliography}{100}
\providecommand{\url}[1]{#1}
\csname url@samestyle\endcsname
\providecommand{\newblock}{\relax}
\providecommand{\bibinfo}[2]{#2}
\providecommand{\BIBentrySTDinterwordspacing}{\spaceskip=0pt\relax}
\providecommand{\BIBentryALTinterwordstretchfactor}{4}
\providecommand{\BIBentryALTinterwordspacing}{\spaceskip=\fontdimen2\font plus
\BIBentryALTinterwordstretchfactor\fontdimen3\font minus
  \fontdimen4\font\relax}
\providecommand{\BIBforeignlanguage}[2]{{%
\expandafter\ifx\csname l@#1\endcsname\relax
\typeout{** WARNING: IEEEtran.bst: No hyphenation pattern has been}%
\typeout{** loaded for the language `#1'. Using the pattern for}%
\typeout{** the default language instead.}%
\else
\language=\csname l@#1\endcsname
\fi
#2}}
\providecommand{\BIBdecl}{\relax}
\BIBdecl

\bibitem{Graphdata}
C.~C. Aggarwal, H.~Wang \emph{et~al.}, \emph{Managing and mining graph
  data}.\hskip 1em plus 0.5em minus 0.4em\relax Springer, 2010, vol.~40.

\bibitem{UserRS}
W.~Fan, Y.~Ma, Q.~Li, Y.~He, E.~Zhao, J.~Tang, and D.~Yin, ``Graph neural
  networks for social recommendation,'' in \emph{The world wide web
  conference}, 2019, pp. 417--426.

\bibitem{Drugdiscovery}
T.~Zhao, Y.~Hu, L.~R. Valsdottir, T.~Zang, and J.~Peng, ``Identifying
  drug--target interactions based on graph convolutional network and deep
  neural network,'' \emph{Briefings in bioinformatics}, vol.~22, no.~2, pp.
  2141--2150, 2021.

\bibitem{Trafficforecast}
Z.~Cui, K.~Henrickson, R.~Ke, and Y.~Wang, ``Traffic graph convolutional
  recurrent neural network: A deep learning framework for network-scale traffic
  learning and forecasting,'' \emph{IEEE Transactions on Intelligent
  Transportation Systems}, vol.~21, no.~11, pp. 4883--4894, 2019.

\bibitem{Pointcloud}
W.~Shi and R.~Rajkumar, ``Point-gnn: Graph neural network for 3d object
  detection in a point cloud,'' in \emph{Proceedings of the IEEE/CVF conference
  on computer vision and pattern recognition}, 2020, pp. 1711--1719.

\bibitem{Chipdesign}
A.~Mirhoseini, A.~Goldie, M.~Yazgan, J.~W. Jiang, E.~Songhori, S.~Wang, Y.-J.
  Lee, E.~Johnson, O.~Pathak, A.~Nazi \emph{et~al.}, ``A graph placement
  methodology for fast chip design,'' \emph{Nature}, vol. 594, no. 7862, pp.
  207--212, 2021.

\bibitem{CNN}
Y.~LeCun, L.~Bottou, Y.~Bengio, and P.~Haffner, ``Gradient-based learning
  applied to document recognition,'' \emph{Proceedings of the IEEE}, vol.~86,
  no.~11, pp. 2278--2324, 1998.

\bibitem{GNN}
F.~Scarselli, M.~Gori, A.~C. Tsoi, M.~Hagenbuchner, and G.~Monfardini, ``The
  graph neural network model,'' \emph{IEEE transactions on neural networks},
  vol.~20, no.~1, pp. 61--80, 2008.

\bibitem{Nodeclf}
K.~Oono and T.~Suzuki, ``Graph neural networks exponentially lose expressive
  power for node classification,'' in \emph{International Conference on
  Learning Representations}.

\bibitem{LinkPred}
M.~Zhang and Y.~Chen, ``Link prediction based on graph neural networks,''
  \emph{Advances in neural information processing systems}, vol.~31, 2018.

\bibitem{Graphclf}
F.~Errica, M.~Podda, D.~Bacciu, and A.~Micheli, ``A fair comparison of graph
  neural networks for graph classification,'' in \emph{International Conference
  on Learning Representations}.

\bibitem{KD}
G.~Hinton, O.~Vinyals, and J.~Dean, ``Distilling the knowledge in a neural
  network,'' \emph{arXiv preprint arXiv:1503.02531}, 2015.

\bibitem{KDincv}
G.~Chen, W.~Choi, X.~Yu, T.~Han, and M.~Chandraker, ``Learning efficient object
  detection models with knowledge distillation,'' \emph{Advances in neural
  information processing systems}, vol.~30, 2017.

\bibitem{Kdinspeech}
Y.~Chebotar and A.~Waters, ``Distilling knowledge from ensembles of neural
  networks for speech recognition.'' in \emph{Interspeech}, 2016, pp.
  3439--3443.

\bibitem{Kdinnlp}
X.~Liu, P.~He, W.~Chen, and J.~Gao, ``Improving multi-task deep neural networks
  via knowledge distillation for natural language understanding,'' \emph{arXiv
  preprint arXiv:1904.09482}, 2019.

\bibitem{LSP}
Y.~Yang, J.~Qiu, M.~Song, D.~Tao, and X.~Wang, ``Distilling knowledge from
  graph convolutional networks,'' in \emph{Proceedings of the IEEE/CVF
  Conference on Computer Vision and Pattern Recognition}, 2020, pp. 7074--7083.

\bibitem{GCN}
T.~N. Kipf and M.~Welling, ``Semi-supervised classification with graph
  convolutional networks,'' \emph{arXiv preprint arXiv:1609.02907}, 2016.

\bibitem{DKWISL}
Z.~Zhang, X.~Shu, B.~Yu, T.~Liu, J.~Zhao, Q.~Li, and L.~Guo, ``Distilling
  knowledge from well-informed soft labels for neural relation extraction,'' in
  \emph{Proceedings of the AAAI Conference on Artificial Intelligence},
  vol.~34, no.~05, 2020, pp. 9620--9627.

\bibitem{KTG}
S.~Minami, T.~Hirakawa, T.~Yamashita, and H.~Fujiyoshi, ``Knowledge transfer
  graph for deep collaborative learning,'' in \emph{Proceedings of the Asian
  Conference on Computer Vision}, 2020.

\bibitem{DGCN}
H.~Wang, D.~Lian, and Y.~Ge, ``Binarized collaborative filtering with
  distilling graph convolutional networks,'' in \emph{Proceedings of the 28th
  International Joint Conference on Artificial Intelligence}, 2019, pp.
  4802--4808.

\bibitem{SPG}
B.~Pan, H.~Cai, D.-A. Huang, K.-H. Lee, A.~Gaidon, E.~Adeli, and J.~C. Niebles,
  ``Spatio-temporal graph for video captioning with knowledge distillation,''
  in \emph{Proceedings of the IEEE/CVF Conference on Computer Vision and
  Pattern Recognition}, 2020, pp. 10\,870--10\,879.

\bibitem{GCLN}
T.~Koji and T.~Kanji, ``Dark reciprocal-rank: Teacher-to-student knowledge
  transfer from self-localization model to graph-convolutional neural
  network,'' in \emph{2021 IEEE International Conference on Robotics and
  Automation (ICRA)}.\hskip 1em plus 0.5em minus 0.4em\relax IEEE, 2021, pp.
  1846--1853.

\bibitem{IEP}
S.~Lee and B.~C. Song, ``Interpretable embedding procedure knowledge transfer
  via stacked principal component analysis and graph neural network,'' in
  \emph{Proceedings of the AAAI Conference on Artificial Intelligence},
  vol.~35, no.~9, 2021, pp. 8297--8305.

\bibitem{HKD}
S.~Zhou, Y.~Wang, D.~Chen, J.~Chen, X.~Wang, C.~Wang, and J.~Bu, ``Distilling
  holistic knowledge with graph neural networks,'' in \emph{Proceedings of the
  IEEE/CVF international conference on computer vision}, 2021, pp.
  10\,387--10\,396.

\bibitem{MHGD}
S.~Lee and B.~C. Song, ``Graph-based knowledge distillation by multi-head
  attention network.''

\bibitem{IRG}
Y.~Liu, J.~Cao, B.~Li, C.~Yuan, W.~Hu, Y.~Li, and Y.~Duan, ``Knowledge
  distillation via instance relationship graph,'' in \emph{Proceedings of the
  IEEE/CVF Conference on Computer Vision and Pattern Recognition}, 2019, pp.
  7096--7104.

\bibitem{DOD}
Y.~Chen, P.~Chen, S.~Liu, L.~Wang, and J.~Jia, ``Deep structured instance graph
  for distilling object detectors,'' in \emph{Proceedings of the IEEE/CVF
  International Conference on Computer Vision}, 2021, pp. 4359--4368.

\bibitem{HKDIFM}
N.~Passalis, M.~Tzelepi, and A.~Tefas, ``Heterogeneous knowledge distillation
  using information flow modeling,'' in \emph{Proceedings of the IEEE/CVF
  Conference on Computer Vision and Pattern Recognition}, 2020, pp. 2339--2348.

\bibitem{KDExplainer}
M.~Xue, J.~Song, X.~Wang, Y.~Chen, X.~Wang, and M.~Song, ``Kdexplainer: A
  task-oriented attention model for explaining knowledge distillation.''

\bibitem{TDD}
J.~Song, H.~Zhang, X.~Wang, M.~Xue, Y.~Chen, L.~Sun, D.~Tao, and M.~Song,
  ``Tree-like decision distillation,'' in \emph{Proceedings of the IEEE/CVF
  Conference on Computer Vision and Pattern Recognition}, 2021, pp.
  13\,488--13\,497.

\bibitem{DualDE}
Y.~Zhu, W.~Zhang, M.~Chen, H.~Chen, X.~Cheng, W.~Zhang, and H.~Chen, ``Dualde:
  Dually distilling knowledge graph embedding for faster and cheaper
  reasoning,'' in \emph{Proceedings of the Fifteenth ACM International
  Conference on Web Search and Data Mining}, 2022, pp. 1516--1524.

\bibitem{CAG}
D.~Guo, H.~Wang, and M.~Wang, ``Context-aware graph inference with knowledge
  distillation for visual dialog,'' \emph{IEEE Transactions on Pattern Analysis
  and Machine Intelligence}, vol.~44, no.~10, pp. 6056--6073, 2021.

\bibitem{GKD}
M.~Ghorbani, M.~Bahrami, A.~Kazi, M.~Soleymani~Baghshah, H.~R. Rabiee, and
  N.~Navab, ``Gkd: Semi-supervised graph knowledge distillation for
  graph-independent inference,'' in \emph{Medical Image Computing and Computer
  Assisted Intervention--MICCAI 2021: 24th International Conference,
  Strasbourg, France, September 27--October 1, 2021, Proceedings, Part V
  24}.\hskip 1em plus 0.5em minus 0.4em\relax Springer, 2021, pp. 709--718.

\bibitem{MorsE}
M.~Chen, W.~Zhang, Y.~Zhu, H.~Zhou, Z.~Yuan, C.~Xu, and H.~Chen,
  ``Meta-knowledge transfer for inductive knowledge graph embedding,'' in
  \emph{Proceedings of the 45th International ACM SIGIR Conference on Research
  and Development in Information Retrieval}, 2022, pp. 927--937.

\bibitem{BAF}
C.~Zhang and Y.~Peng, ``Better and faster: knowledge transfer from multiple
  self-supervised learning tasks via graph distillation for video
  classification,'' in \emph{Proceedings of the 27th International Joint
  Conference on Artificial Intelligence}, 2018, pp. 1135--1141.

\bibitem{LAD}
T.~He, J.~Chen, X.~Tan, and T.~Qin, ``Language graph distillation for
  low-resource machine translation,'' \emph{arXiv preprint arXiv:1908.06258},
  2019.

\bibitem{GD}
Z.~Luo, J.-T. Hsieh, L.~Jiang, J.~C. Niebles, and L.~Fei-Fei, ``Graph
  distillation for action detection with privileged modalities,'' in
  \emph{Proceedings of the European Conference on Computer Vision (ECCV)},
  2018, pp. 166--183.

\bibitem{GCMT}
X.~Liu and S.~Zhang, ``Graph consistency based mean-teaching for unsupervised
  domain adaptive person re-identification.''

\bibitem{GraSSNet}
Y.~Zhang, M.~Jiang, and Q.~Zhao, ``Saliency prediction with external
  knowledge,'' in \emph{Proceedings of the IEEE/CVF winter conference on
  applications of computer vision}, 2021, pp. 484--493.

\bibitem{LSN}
H.~Chen, Y.~Wang, C.~Xu, C.~Xu, and D.~Tao, ``Learning student networks via
  feature embedding,'' \emph{IEEE Transactions on Neural Networks and Learning
  Systems}, vol.~32, no.~1, pp. 25--35, 2020.

\bibitem{IntRA-KD}
Y.~Hou, Z.~Ma, C.~Liu, T.-W. Hui, and C.~C. Loy, ``Inter-region affinity
  distillation for road marking segmentation,'' in \emph{Proceedings of the
  IEEE/CVF Conference on Computer Vision and Pattern Recognition}, 2020, pp.
  12\,486--12\,495.

\bibitem{RKD}
W.~Park, D.~Kim, Y.~Lu, and M.~Cho, ``Relational knowledge distillation,'' in
  \emph{Proceedings of the IEEE/CVF Conference on Computer Vision and Pattern
  Recognition}, 2019, pp. 3967--3976.

\bibitem{CC}
B.~Peng, X.~Jin, J.~Liu, D.~Li, Y.~Wu, Y.~Liu, S.~Zhou, and Z.~Zhang,
  ``Correlation congruence for knowledge distillation,'' in \emph{Proceedings
  of the IEEE/CVF International Conference on Computer Vision}, 2019, pp.
  5007--5016.

\bibitem{SPKD}
F.~Tung and G.~Mori, ``Similarity-preserving knowledge distillation,'' in
  \emph{Proceedings of the IEEE/CVF International Conference on Computer
  Vision}, 2019, pp. 1365--1374.

\bibitem{KCAN}
K.~Tu, P.~Cui, D.~Wang, Z.~Zhang, J.~Zhou, Y.~Qi, and W.~Zhu, ``Conditional
  graph attention networks for distilling and refining knowledge graphs in
  recommendation,'' in \emph{Proceedings of the 30th ACM International
  Conference on Information \& Knowledge Management}, 2021, pp. 1834--1843.

\bibitem{GFKD}
X.~Deng and Z.~Zhang, ``Graph-free knowledge distillation for graph neural
  networks.''

\bibitem{RDD}
W.~Zhang, X.~Miao, Y.~Shao, J.~Jiang, L.~Chen, O.~Ruas, and B.~Cui, ``Reliable
  data distillation on graph convolutional network,'' in \emph{Proceedings of
  the 2020 ACM SIGMOD International Conference on Management of Data}, 2020,
  pp. 1399--1414.

\bibitem{GKD2}
C.~Lassance, M.~Bontonou, G.~B. Hacene, V.~Gripon, J.~Tang, and A.~Ortega,
  ``Deep geometric knowledge distillation with graphs,'' in \emph{ICASSP
  2020-2020 IEEE International Conference on Acoustics, Speech and Signal
  Processing (ICASSP)}.\hskip 1em plus 0.5em minus 0.4em\relax IEEE, 2020, pp.
  8484--8488.

\bibitem{GLNN}
S.~Zhang, Y.~Liu, Y.~Sun, and N.~Shah, ``Graph-less neural networks: Teaching
  old mlps new tricks via distillation,'' in \emph{International Conference on
  Learning Representations}.

\bibitem{Distill2Vec}
S.~Antaris and D.~Rafailidis, ``Distill2vec: dynamic graph representation
  learning with knowledge distillation,'' in \emph{2020 IEEE/ACM International
  Conference on Advances in Social Networks Analysis and Mining
  (ASONAM)}.\hskip 1em plus 0.5em minus 0.4em\relax IEEE, 2020, pp. 60--64.

\bibitem{MT-GCN}
K.~Zhan and C.~Niu, ``Mutual teaching for graph convolutional networks,''
  \emph{Future Generation Computer Systems}, vol. 115, pp. 837--843, 2021.

\bibitem{TinyGNN}
B.~Yan, C.~Wang, G.~Guo, and Y.~Lou, ``Tinygnn: Learning efficient graph neural
  networks,'' in \emph{Proceedings of the 26th ACM SIGKDD International
  Conference on Knowledge Discovery \& Data Mining}, 2020, pp. 1848--1856.

\bibitem{GLocalKD}
R.~Ma, G.~Pang, L.~Chen, and A.~van~den Hengel, ``Deep graph-level anomaly
  detection by glocal knowledge distillation,'' in \emph{Proceedings of the
  Fifteenth ACM International Conference on Web Search and Data Mining}, 2022,
  pp. 704--714.

\bibitem{SCR}
C.~Zhang, Y.~He, Y.~Cen, Z.~Hou, W.~Feng, Y.~Dong, X.~Cheng, H.~Cai, F.~He, and
  J.~Tang, ``Scr: Training graph neural networks with consistency
  regularization,'' \emph{arXiv e-prints}, pp. arXiv--2112, 2021.

\bibitem{ROD}
W.~Zhang, Y.~Jiang, Y.~Li, Z.~Sheng, Y.~Shen, X.~Miao, L.~Wang, Z.~Yang, and
  B.~Cui, ``Rod: reception-aware online distillation for sparse graphs,'' in
  \emph{Proceedings of the 27th ACM SIGKDD Conference on Knowledge Discovery \&
  Data Mining}, 2021, pp. 2232--2242.

\bibitem{EGNN}
Y.~Li, L.~Liu, G.~Wang, Y.~Du, and P.~Chen, ``Egnn: Constructing explainable
  graph neural networks via knowledge distillation,'' \emph{Knowledge-Based
  Systems}, vol. 241, p. 108345, 2022.

\bibitem{LWC-KD}
Y.~Wang, Y.~Zhang, and M.~Coates, ``Graph structure aware contrastive knowledge
  distillation for incremental learning in recommender systems,'' in
  \emph{Proceedings of the 30th ACM International Conference on Information \&
  Knowledge Management}, 2021, pp. 3518--3522.

\bibitem{MustaD}
J.~Kim, J.~Jung, and U.~Kang, ``Compressing deep graph convolution network with
  multi-staged knowledge distillation,'' \emph{Plos one}, vol.~16, no.~8, p.
  e0256187, 2021.

\bibitem{EGAD}
S.~Antaris, D.~Rafailidis, and S.~Girdzijauskas, ``Egad: Evolving graph
  representation learning with self-attention and knowledge distillation for
  live video streaming events,'' in \emph{2020 IEEE International Conference on
  Big Data (Big Data)}.\hskip 1em plus 0.5em minus 0.4em\relax IEEE, 2020, pp.
  1455--1464.

\bibitem{AGNN}
Y.~Jing, Y.~Yang, X.~Wang, M.~Song, and D.~Tao, ``Amalgamating knowledge from
  heterogeneous graph neural networks,'' in \emph{Proceedings of the IEEE/CVF
  conference on computer vision and pattern recognition}, 2021, pp.
  15\,709--15\,718.

\bibitem{ColdBrew}
W.~Zheng, E.~W. Huang, N.~Rao, S.~Katariya, Z.~Wang, and K.~Subbian, ``Cold
  brew: Distilling graph node representations with incomplete or missing
  neighborhoods,'' \emph{arXiv preprint arXiv:2111.04840}, 2021.

\bibitem{PGD}
S.~Wang, K.~Zhang, L.~Wu, H.~Ma, R.~Hong, and M.~Wang, ``Privileged graph
  distillation for cold start recommendation,'' in \emph{Proceedings of the
  44th International ACM SIGIR Conference on Research and Development in
  Information Retrieval}, 2021, pp. 1187--1196.

\bibitem{OAD}
C.~Wang, Z.~Wang, D.~Chen, S.~Zhou, Y.~Feng, and C.~Chen, ``Online adversarial
  distillation for graph neural networks,'' \emph{arXiv preprint
  arXiv:2112.13966}, 2021.

\bibitem{CKD}
C.~Wang, S.~Zhou, K.~Yu, D.~Chen, B.~Li, Y.~Feng, and C.~Chen, ``Collaborative
  knowledge distillation for heterogeneous information network embedding,'' in
  \emph{Proceedings of the ACM Web Conference 2022}, 2022, pp. 1631--1639.

\bibitem{BGNN}
M.~Bahri, G.~Bahl, and S.~Zafeiriou, ``Binary graph neural networks,'' in
  \emph{Proceedings of the IEEE/CVF Conference on Computer Vision and Pattern
  Recognition}, 2021, pp. 9492--9501.

\bibitem{EGSC}
C.~Qin, H.~Zhao, L.~Wang, H.~Wang, Y.~Zhang, and Y.~Fu, ``Slow learning and
  fast inference: Efficient graph similarity computation via knowledge
  distillation,'' \emph{Advances in Neural Information Processing Systems},
  vol.~34, pp. 14\,110--14\,121, 2021.

\bibitem{HSKDM}
Z.~Huang, Y.~Tang, and Y.~Chen, ``A graph neural network-based node
  classification model on class-imbalanced graph data,'' \emph{Knowledge-Based
  Systems}, vol. 244, p. 108538, 2022.

\bibitem{GRL}
J.~Ma and Q.~Mei, ``Graph representation learning via multi-task knowledge
  distillation,'' in \emph{33rd Conference on Neural Information Processing
  Systems (NeurIPS 2019) Graph Representation Learning Workshop}, 2019.

\bibitem{GFL}
H.~Yao, C.~Zhang, Y.~Wei, M.~Jiang, S.~Wang, J.~Huang, N.~Chawla, and Z.~Li,
  ``Graph few-shot learning via knowledge transfer,'' in \emph{Proceedings of
  the AAAI Conference on Artificial Intelligence}, vol.~34, no.~04, 2020, pp.
  6656--6663.

\bibitem{HGKT}
J.~Wang, X.~Wang, B.~Jin, J.~Yan, W.~Zhang, and H.~Zha, ``Heterogeneous
  graph-based knowledge transfer for generalized zero-shot learning,'' in
  \emph{2020 25th International Conference on Pattern Recognition
  (ICPR)}.\hskip 1em plus 0.5em minus 0.4em\relax IEEE, 2021, pp. 1859--1866.

\bibitem{CPF}
C.~Yang, J.~Liu, and C.~Shi, ``Extract the knowledge of graph neural networks
  and go beyond it: An effective knowledge distillation framework,'' in
  \emph{Proceedings of the web conference 2021}, 2021, pp. 1227--1237.

\bibitem{scGCN}
Q.~Song, J.~Su, and W.~Zhang, ``scgcn is a graph convolutional networks
  algorithm for knowledge transfer in single cell omics,'' \emph{Nature
  communications}, vol.~12, no.~1, p. 3826, 2021.

\bibitem{MetaHG}
Y.~Qian, Y.~Zhang, Y.~Ye, C.~Zhang \emph{et~al.}, ``Distilling meta knowledge
  on heterogeneous graph for illicit drug trafficker detection on social
  media,'' \emph{Advances in Neural Information Processing Systems}, vol.~34,
  pp. 26\,911--26\,923, 2021.

\bibitem{G-CRD}
C.~K. Joshi, F.~Liu, X.~Xun, J.~Lin, and C.~S. Foo, ``On representation
  knowledge distillation for graph neural networks,'' \emph{IEEE Transactions
  on Neural Networks and Learning Systems}, 2022.

\bibitem{HIRE}
J.~Liu, T.~Zheng, and Q.~Hao, ``Hire: Distilling high-order relational
  knowledge from heterogeneous graph neural networks,'' \emph{Neurocomputing},
  vol. 507, pp. 67--83, 2022.

\bibitem{LinkDist}
Y.~Luo, A.~Chen, K.~Yan, and L.~Tian, ``Distilling self-knowledge from
  contrastive links to classify graph nodes without passing messages,''
  \emph{arXiv preprint arXiv:2106.08541}, 2021.

\bibitem{IGSD}
H.~Zhang, S.~Lin, W.~Liu, P.~Zhou, J.~Tang, X.~Liang, and E.~P. Xing,
  ``Iterative graph self-distillation,'' \emph{arXiv preprint
  arXiv:2010.12609}, 2020.

\bibitem{SAIL}
L.~Yu, S.~Pei, L.~Ding, J.~Zhou, L.~Li, C.~Zhang, and X.~Zhang, ``Sail:
  Self-augmented graph contrastive learning,'' in \emph{Proceedings of the AAAI
  Conference on Artificial Intelligence}, vol.~36, no.~8, 2022, pp. 8927--8935.

\bibitem{GNN-SD}
Y.~Chen, Y.~Bian, X.~Xiao, Y.~Rong, T.~Xu, and J.~Huang, ``On self-distilling
  graph neural network.''

\bibitem{SDSS}
Y.~Ren, J.~Ji, L.~Niu, and M.~Lei, ``Multi-task self-distillation for
  graph-based semi-supervised learning,'' \emph{arXiv preprint
  arXiv:2112.01174}, 2021.

\bibitem{MEIRec}
S.~Fan, J.~Zhu, X.~Han, C.~Shi, L.~Hu, B.~Ma, and Y.~Li, ``Metapath-guided
  heterogeneous graph neural network for intent recommendation,'' in
  \emph{Proceedings of the 25th ACM SIGKDD international conference on
  knowledge discovery \& data mining}, 2019, pp. 2478--2486.

\bibitem{Chemical}
F.~Wang, J.-F. Yang, M.-Y. Wang, C.-Y. Jia, X.-X. Shi, G.-F. Hao, and G.-F.
  Yang, ``Graph attention convolutional neural network model for chemical
  poisoning of honey bees’ prediction,'' \emph{Science Bulletin}, vol.~65,
  no.~14, pp. 1184--1191, 2020.

\bibitem{Physics}
P.~Battaglia, R.~Pascanu, M.~Lai, D.~Jimenez~Rezende \emph{et~al.},
  ``Interaction networks for learning about objects, relations and physics,''
  \emph{Advances in neural information processing systems}, vol.~29, 2016.

\bibitem{KG}
X.~Lin, Z.~Quan, Z.-J. Wang, T.~Ma, and X.~Zeng, ``Kgnn: knowledge graph neural
  network for drug-drug interaction prediction,'' in \emph{Proceedings of the
  Twenty-Ninth International Conference on International Joint Conferences on
  Artificial Intelligence}, 2021, pp. 2739--2745.

\bibitem{Circuit}
G.~Zhang, H.~He, and D.~Katabi, ``Circuit-gnn: Graph neural networks for
  distributed circuit design,'' in \emph{International conference on machine
  learning}.\hskip 1em plus 0.5em minus 0.4em\relax PMLR, 2019, pp. 7364--7373.

\bibitem{SpectralGNN}
J.~Bruna, W.~Zaremba, A.~Szlam, and Y.~LeCun, ``Spectral networks and deep
  locally connected networks on graphs,'' in \emph{2nd International Conference
  on Learning Representations, ICLR}, 2014.

\bibitem{Spectraltheory}
F.~R. Chung, \emph{Spectral graph theory}.\hskip 1em plus 0.5em minus
  0.4em\relax American Mathematical Soc., 1997, vol.~92.

\bibitem{ChebyNet}
M.~Defferrard, X.~Bresson, and P.~Vandergheynst, ``Convolutional neural
  networks on graphs with fast localized spectral filtering,'' \emph{Advances
  in neural information processing systems}, vol.~29, 2016.

\bibitem{AGCN}
R.~Li, S.~Wang, F.~Zhu, and J.~Huang, ``Adaptive graph convolutional neural
  networks,'' in \emph{Proceedings of the AAAI conference on artificial
  intelligence}, vol.~32, no.~1, 2018.

\bibitem{zhuang2018dual}
C.~Zhuang and Q.~Ma, ``Dual graph convolutional networks for graph-based
  semi-supervised classification,'' in \emph{Proceedings of the 2018 world wide
  web conference}, 2018, pp. 499--508.

\bibitem{xugraph}
B.~Xu, H.~Shen, Q.~Cao, Y.~Qiu, and X.~Cheng, ``Graph wavelet neural network,''
  in \emph{International Conference on Learning Representations}.

\bibitem{SAGE}
W.~Hamilton, Z.~Ying, and J.~Leskovec, ``Inductive representation learning on
  large graphs,'' \emph{Advances in neural information processing systems},
  vol.~30, 2017.

\bibitem{FastGCN}
J.~Chen, T.~Ma, and C.~Xiao, ``Fastgcn: Fast learning with graph convolu-tional
  networks via importance sampling,'' in \emph{International Conference on
  Learning Representations}.\hskip 1em plus 0.5em minus 0.4em\relax
  International Conference on Learning Representations, ICLR, 2018.

\bibitem{LADIES}
D.~Zou, Z.~Hu, Y.~Wang, S.~Jiang, Y.~Sun, and Q.~Gu, ``Layer-dependent
  importance sampling for training deep and large graph convolutional
  networks,'' \emph{Advances in neural information processing systems},
  vol.~32, 2019.

\bibitem{GAT}
P.~Veli{\v{c}}kovi{\'c}, G.~Cucurull, A.~Casanova, A.~Romero, P.~Li{\`o}, and
  Y.~Bengio, ``Graph attention networks,'' in \emph{International Conference on
  Learning Representations}.

\bibitem{atwood2016diffusion}
J.~Atwood and D.~Towsley, ``Diffusion-convolutional neural networks,''
  \emph{Advances in neural information processing systems}, vol.~29, 2016.

\bibitem{MPNN}
J.~Gilmer, S.~S. Schoenholz, P.~F. Riley, O.~Vinyals, and G.~E. Dahl, ``Neural
  message passing for quantum chemistry,'' in \emph{International conference on
  machine learning}.\hskip 1em plus 0.5em minus 0.4em\relax PMLR, 2017, pp.
  1263--1272.

\bibitem{wang2018non}
X.~Wang, R.~Girshick, A.~Gupta, and K.~He, ``Non-local neural networks,'' in
  \emph{Proceedings of the IEEE conference on computer vision and pattern
  recognition}, 2018, pp. 7794--7803.

\bibitem{RGCN}
P.~W. Battaglia, J.~B. Hamrick, V.~Bapst, A.~Sanchez-Gonzalez, V.~Zambaldi,
  M.~Malinowski, A.~Tacchetti, D.~Raposo, A.~Santoro, R.~Faulkner
  \emph{et~al.}, ``Relational inductive biases, deep learning, and graph
  networks,'' \emph{arXiv preprint arXiv:1806.01261}, 2018.

\bibitem{zhou2020graph}
J.~Zhou, G.~Cui, S.~Hu, Z.~Zhang, C.~Yang, Z.~Liu, L.~Wang, C.~Li, and M.~Sun,
  ``Graph neural networks: A review of methods and applications,'' \emph{AI
  open}, vol.~1, pp. 57--81, 2020.

\bibitem{wu2020comprehensive}
Z.~Wu, S.~Pan, F.~Chen, G.~Long, C.~Zhang, and S.~Y. Philip, ``A comprehensive
  survey on graph neural networks,'' \emph{IEEE transactions on neural networks
  and learning systems}, vol.~32, no.~1, pp. 4--24, 2020.

\bibitem{zhang2020deep}
Z.~Zhang, P.~Cui, and W.~Zhu, ``Deep learning on graphs: A survey,'' \emph{IEEE
  Transactions on Knowledge and Data Engineering}, vol.~34, no.~1, pp.
  249--270, 2020.

\bibitem{skarding2021foundations}
J.~Skarding, B.~Gabrys, and K.~Musial, ``Foundations and modeling of dynamic
  networks using dynamic graph neural networks: A survey,'' \emph{IEEE Access},
  vol.~9, pp. 79\,143--79\,168, 2021.

\bibitem{yang2020heterogeneous}
C.~Yang, Y.~Xiao, Y.~Zhang, Y.~Sun, and J.~Han, ``Heterogeneous network
  representation learning: A unified framework with survey and benchmark,''
  \emph{IEEE Transactions on Knowledge and Data Engineering}, vol.~34, no.~10,
  pp. 4854--4873, 2020.

\bibitem{sun2018adversarial}
L.~Sun, Y.~Dou, C.~Yang, J.~Wang, P.~S. Yu, L.~He, and B.~Li, ``Adversarial
  attack and defense on graph data: A survey,'' \emph{arXiv preprint
  arXiv:1812.10528}, 2018.

\bibitem{wu2023graph}
L.~Wu, Y.~Chen, K.~Shen, X.~Guo, H.~Gao, S.~Li, J.~Pei, B.~Long \emph{et~al.},
  ``Graph neural networks for natural language processing: A survey,''
  \emph{Foundations and Trends{\textregistered} in Machine Learning}, vol.~16,
  no.~2, pp. 119--328, 2023.

\bibitem{nazir2021survey}
U.~Nazir, H.~Wang, and M.~Taj, ``Survey of image based graph neural networks,''
  \emph{arXiv preprint arXiv:2106.06307}, 2021.

\bibitem{lopera2021survey}
D.~S. Lopera, L.~Servadei, G.~N. Kiprit, S.~Hazra, R.~Wille, and W.~Ecker, ``A
  survey of graph neural networks for electronic design automation,'' in
  \emph{2021 ACM/IEEE 3rd Workshop on Machine Learning for CAD (MLCAD)}.\hskip
  1em plus 0.5em minus 0.4em\relax IEEE, 2021, pp. 1--6.

\bibitem{wu2022graph}
S.~Wu, F.~Sun, W.~Zhang, X.~Xie, and B.~Cui, ``Graph neural networks in
  recommender systems: a survey,'' \emph{ACM Computing Surveys}, vol.~55,
  no.~5, pp. 1--37, 2022.

\bibitem{lamb2020graph}
L.~C. Lamb, A.~Garcez, M.~Gori, M.~Prates, P.~Avelar, and M.~Vardi, ``Graph
  neural networks meet neural-symbolic computing: A survey and perspective,''
  in \emph{IJCAI-PRICAI 2020-29th International Joint Conference on Artificial
  Intelligence-Pacific Rim International Conference on Artificial
  Intelligence}, 2020.

\bibitem{Labelsmooth}
L.~Yuan, F.~E. Tay, G.~Li, T.~Wang, and J.~Feng, ``Revisiting knowledge
  distillation via label smoothing regularization,'' in \emph{Proceedings of
  the IEEE/CVF Conference on Computer Vision and Pattern Recognition}, 2020,
  pp. 3903--3911.

\bibitem{DML}
Y.~Zhang, T.~Xiang, T.~M. Hospedales, and H.~Lu, ``Deep mutual learning,'' in
  \emph{Proceedings of the IEEE conference on computer vision and pattern
  recognition}, 2018, pp. 4320--4328.

\bibitem{BAN}
T.~Furlanello, Z.~Lipton, M.~Tschannen, L.~Itti, and A.~Anandkumar, ``Born
  again neural networks,'' in \emph{International Conference on Machine
  Learning}.\hskip 1em plus 0.5em minus 0.4em\relax PMLR, 2018, pp. 1607--1616.

\bibitem{FitNet}
A.~Romero, N.~Ballas, S.~E. Kahou, A.~Chassang, C.~Gatta, and Y.~Bengio,
  ``Fitnets: Hints for thin deep nets,'' \emph{arXiv preprint arXiv:1412.6550},
  2014.

\bibitem{AT}
S.~Zagoruyko and N.~Komodakis, ``Paying more attention to attention: Improving
  the performance of convolutional neural networks via attention transfer,'' in
  \emph{International Conference on Learning Representations}.

\bibitem{FT}
J.~Kim, S.~Park, and N.~Kwak, ``Paraphrasing complex network: Network
  compression via factor transfer,'' \emph{Advances in neural information
  processing systems}, vol.~31, 2018.

\bibitem{Alp-kd}
P.~Passban, Y.~Wu, M.~Rezagholizadeh, and Q.~Liu, ``Alp-kd: Attention-based
  layer projection for knowledge distillation,'' in \emph{Proceedings of the
  AAAI Conference on artificial intelligence}, vol.~35, no.~15, 2021, pp.
  13\,657--13\,665.

\bibitem{yim2017gift}
J.~Yim, D.~Joo, J.~Bae, and J.~Kim, ``A gift from knowledge distillation: Fast
  optimization, network minimization and transfer learning,'' in
  \emph{Proceedings of the IEEE conference on computer vision and pattern
  recognition}, 2017, pp. 4133--4141.

\bibitem{passalis2020probabilistic}
N.~Passalis, M.~Tzelepi, and A.~Tefas, ``Probabilistic knowledge transfer for
  lightweight deep representation learning,'' \emph{IEEE Transactions on Neural
  Networks and Learning Systems}, vol.~32, no.~5, pp. 2030--2039, 2020.

\bibitem{CIFAR}
A.~Krizhevsky, ``Learning multiple layers of features from tiny images,''
  \emph{Master's thesis, University of Tront}, 2009.

\bibitem{CoraCiteseerPubmed}
Z.~Yang, W.~Cohen, and R.~Salakhudinov, ``Revisiting semi-supervised learning
  with graph embeddings,'' in \emph{International conference on machine
  learning}.\hskip 1em plus 0.5em minus 0.4em\relax PMLR, 2016, pp. 40--48.

\bibitem{Amazondata}
O.~Shchur, M.~Mumme, A.~Bojchevski, and S.~G{\"u}nnemann, ``Pitfalls of graph
  neural network evaluation,'' \emph{arXiv preprint arXiv:1811.05868}, 2018.

\bibitem{Resnet}
K.~He, X.~Zhang, S.~Ren, and J.~Sun, ``Deep residual learning for image
  recognition,'' in \emph{Proceedings of the IEEE conference on computer vision
  and pattern recognition}, 2016, pp. 770--778.

\bibitem{t-sne}
L.~Van~der Maaten and G.~Hinton, ``Visualizing data using t-sne.''
  \emph{Journal of machine learning research}, vol.~9, no.~11, 2008.

\bibitem{dgl}
M.~Y. Wang, ``Deep graph library: Towards efficient and scalable deep learning
  on graphs,'' in \emph{ICLR workshop on representation learning on graphs and
  manifolds}, 2019.

\bibitem{pytorch}
A.~Paszke, S.~Gross, F.~Massa, A.~Lerer, J.~Bradbury, G.~Chanan, T.~Killeen,
  Z.~Lin, N.~Gimelshein, L.~Antiga \emph{et~al.}, ``Pytorch: An imperative
  style, high-performance deep learning library,'' \emph{Advances in neural
  information processing systems}, vol.~32, 2019.

\bibitem{li2020gan}
M.~Li, J.~Lin, Y.~Ding, Z.~Liu, J.-Y. Zhu, and S.~Han, ``Gan compression:
  Efficient architectures for interactive conditional gans,'' in
  \emph{Proceedings of the IEEE/CVF conference on computer vision and pattern
  recognition}, 2020, pp. 5284--5294.

\bibitem{cheng2020explaining}
X.~Cheng, Z.~Rao, Y.~Chen, and Q.~Zhang, ``Explaining knowledge distillation by
  quantifying the knowledge,'' in \emph{Proceedings of the IEEE/CVF conference
  on computer vision and pattern recognition}, 2020, pp. 12\,925--12\,935.

\bibitem{mobahi2020self}
H.~Mobahi, M.~Farajtabar, and P.~Bartlett, ``Self-distillation amplifies
  regularization in hilbert space,'' \emph{Advances in Neural Information
  Processing Systems}, vol.~33, pp. 3351--3361, 2020.

\bibitem{papernot2016distillation}
N.~Papernot, P.~McDaniel, X.~Wu, S.~Jha, and A.~Swami, ``Distillation as a
  defense to adversarial perturbations against deep neural networks,'' in
  \emph{2016 IEEE symposium on security and privacy (SP)}.\hskip 1em plus 0.5em
  minus 0.4em\relax IEEE, 2016, pp. 582--597.

\bibitem{ross2018improving}
A.~Ross and F.~Doshi-Velez, ``Improving the adversarial robustness and
  interpretability of deep neural networks by regularizing their input
  gradients,'' in \emph{Proceedings of the AAAI Conference on Artificial
  Intelligence}, vol.~32, no.~1, 2018.

\bibitem{papernotsemi}
N.~Papernot, M.~Abadi, {\'U}.~Erlingsson, I.~Goodfellow, and K.~Talwar,
  ``Semi-supervised knowledge transfer for deep learning from private training
  data,'' in \emph{International Conference on Learning Representations}.

\bibitem{wang2019private}
J.~Wang, W.~Bao, L.~Sun, X.~Zhu, B.~Cao, and S.~Y. Philip, ``Private model
  compression via knowledge distillation,'' in \emph{Proceedings of the AAAI
  Conference on Artificial Intelligence}, vol.~33, no.~01, 2019, pp.
  1190--1197.

\end{thebibliography}

\newpage
\begin{IEEEbiography}
    [{\includegraphics[width=1in,height=1.25in,clip,keepaspectratio]{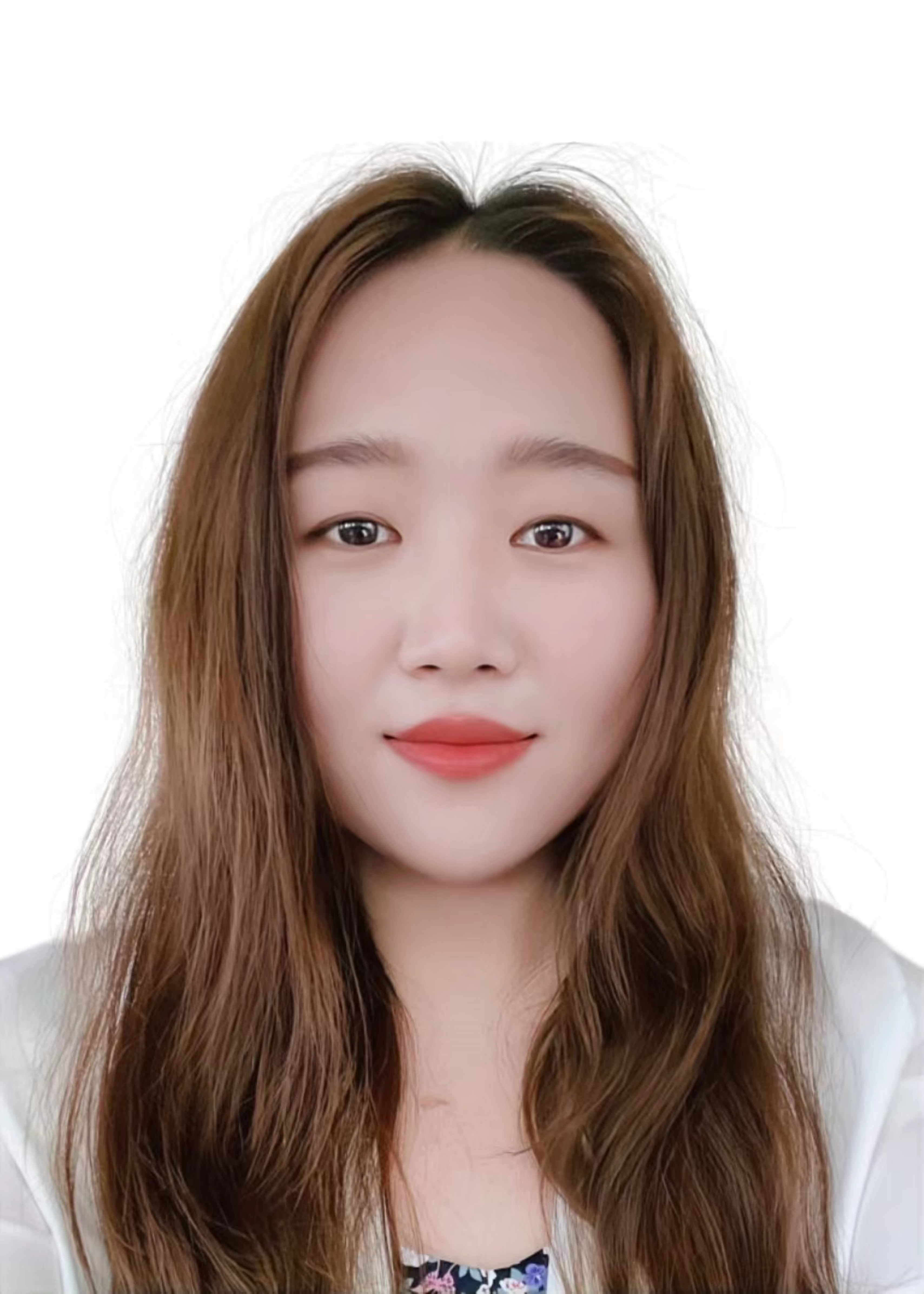}}]
	{Jing Liu}
	is currently pursuing the Ph.D. degree with the Institute of Computing Technology, Chinese Academy of Sciences. Her research interests include graph neural networks, heterogeneous graph representation learning, and knowledge distillation.
\end{IEEEbiography}
\vskip -2\baselineskip plus -1fil

\begin{IEEEbiography}
    [{\includegraphics[width=1in,height=1.25in,clip,keepaspectratio]{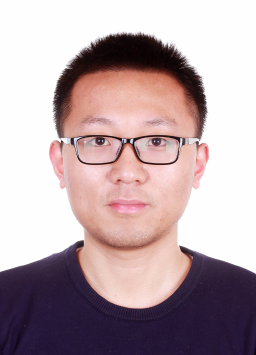}}]
	{Tongya Zheng}
    \zty{is currently pursuing the Ph.D. degree with the College of Computer Science, Zhejiang University. He received his B.Eng. Degree from Nanjing University of Science and Technology. His research interests include graph neural networks, temporal graphs, and explanation for artificial intelligence. He has authored and co-authored many scientific articles at top venues including IEEE TNNLS and AAAI. He has served with international conferences including AAAI and ECML-PKDD, and international journals including Information Sciences.}
\end{IEEEbiography}
\vskip -2\baselineskip plus -1fil

\begin{IEEEbiography}
    [{\includegraphics[width=1in,height=1.25in,clip,keepaspectratio]{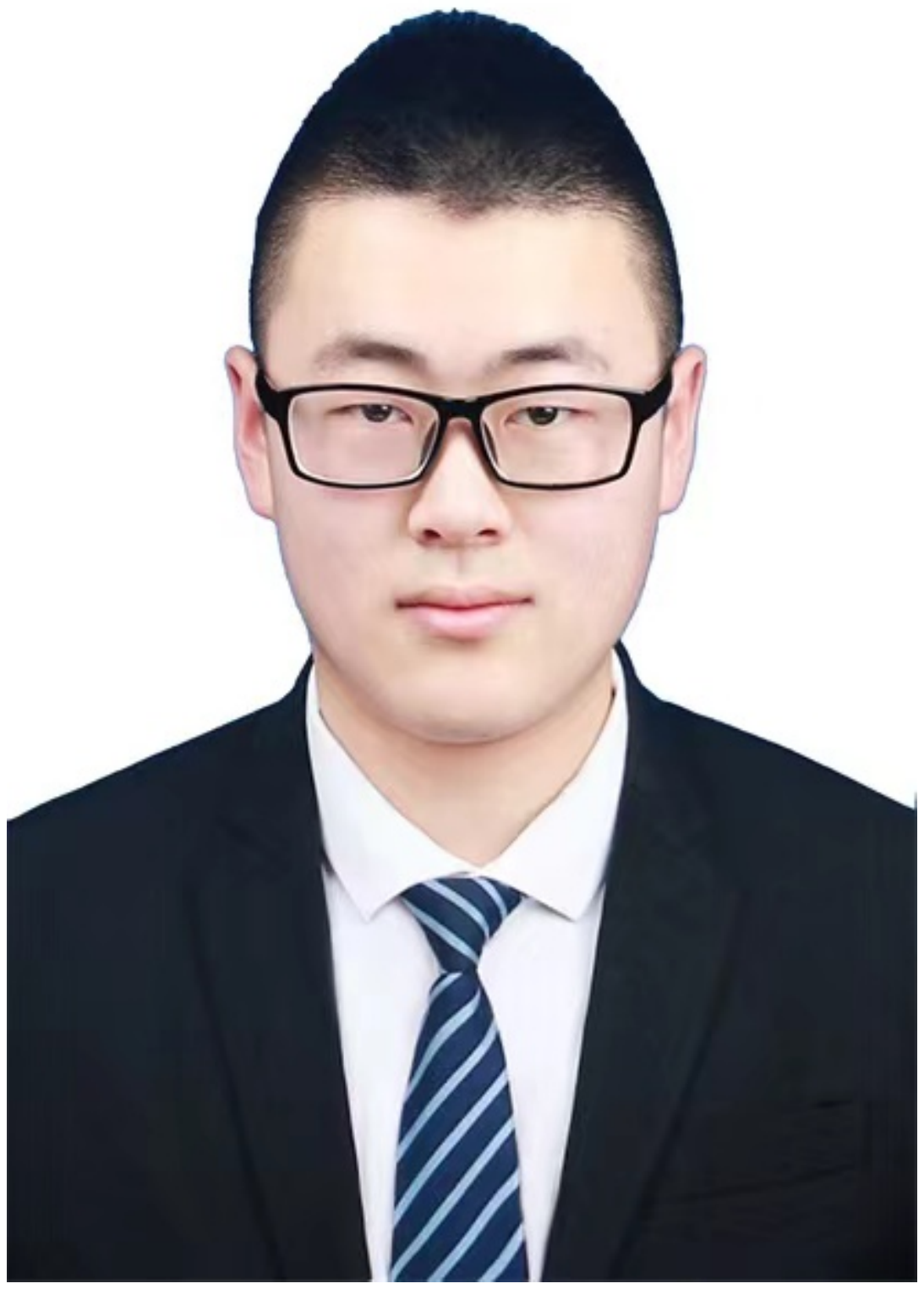}}]
	{Guanzheng Zhang}
	is currently pursuing the M.S. degree with the College of Information Engineering, China University of Geosciences. His research interests include machine learning and graph mining.
\end{IEEEbiography}
\vskip -2\baselineskip plus -1fil

\begin{IEEEbiography}
    [{\includegraphics[width=1in,height=1.25in,clip,keepaspectratio]{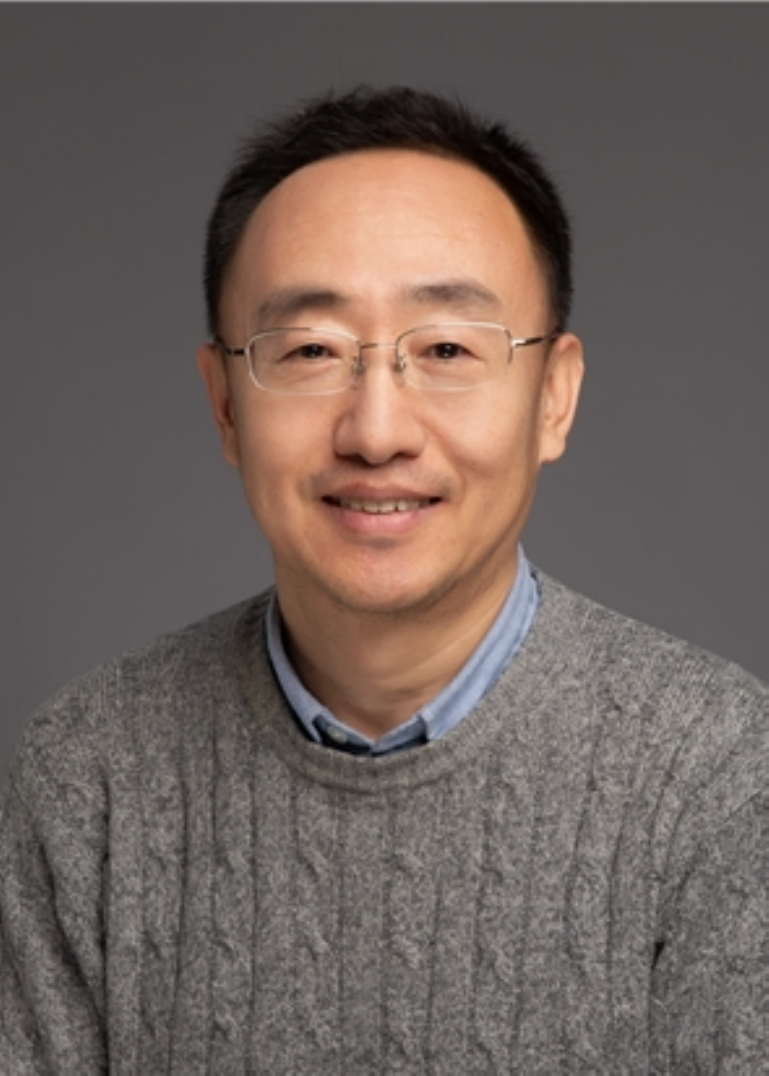}}]
	{Qinfen Hao}
	received the Ph.D. degree in computer system architecture from the Institute of Computing Technology, Chinese Academy of Sciences, China, in 2006. He is currently a Researcher at the Institute of Computing, Chinese Academy of Sciences. His research interests include computer architecture and graph computing.
\end{IEEEbiography}

\end{document}